\documentclass[journal]{IEEEtran}
\usepackage{cite}
\usepackage{hyperref}
\usepackage{amsthm, amssymb, amsmath}
\usepackage{graphicx}
\usepackage{epstopdf}
\usepackage{float}
\usepackage{rotating}
\usepackage{subfigure}
\usepackage[lined,ruled,linesnumbered,commentsnumbered]{algorithm2e}
\usepackage{cleveref}
\usepackage{booktabs}
\usepackage{amsmath,breqn}
\usepackage{multirow}
\usepackage{amsmath}
\usepackage{pdflscape}
\usepackage{color, colortbl}
\usepackage{threeparttable} 
\usepackage{array}
\usepackage[numbers,sort&compress]{natbib}

\begin{document}
\title{An Efficient Dynamic Resource Allocation Framework for Evolutionary Bilevel Optimization}
\author{Dejun~Xu,
       Kai~Ye,
       Zimo~Zheng,
       Tao~Zhou,
       Gary G.~Yen,~\IEEEmembership{Fellow,~IEEE}
       and Min~Jiang,~\IEEEmembership{Senior Member,~IEEE}
\thanks{This work was supported in part by the National Natural Science Foundation of China under Grant No. 62276222, and in part by the Public Technology Service Platform Project of Xiamen City under Grant No.3502Z20231043.
\textit{(Corresponding author: Min Jiang)}
}
\thanks{D. Xu, K. Ye, Z. Zheng, T. Zhou and M. Jiang are with the Department of Artificial Intelligence, School of Informatics, Xiamen University; the Key Laboratory of Multimedia Trusted Perception and Efficient Computing, Ministry of Education; and the Key Laboratory of Digital Protection and Intelligent Processing of Intangible Cultural Heritage of Fujian and Taiwan, Ministry of Culture and Tourism, Xiamen 361005, China (e-mail: xudejun@stu.xmu.edu.cn; minjiang@xmu.edu.cn).}
\thanks{Gary G. Yen is with the School of Electrical and Computer Engineering, Oklahoma State University, Stillwater, OK 74075 USA (e-mail: gyen@okstate.edu).}
}

\markboth{}
{Shell \MakeLowercase{\textit{et al.}}: Bare Demo of IEEEtran.cls for IEEE Journals}
\maketitle

\begin{abstract}
Bilevel optimization problems are characterized by an interactive hierarchical structure, where the upper level seeks to optimize its strategy while simultaneously considering the response of the lower level.
Evolutionary algorithms are commonly used to solve complex bilevel problems in practical scenarios, but they face significant resource consumption challenges due to the nested structure imposed by the implicit lower-level optimality condition. 
This challenge becomes even more pronounced as problem dimensions increase.
Although recent methods have enhanced bilevel convergence through task-level knowledge sharing, further efficiency improvements are still hindered by redundant lower-level iterations that consume excessive resources while generating unpromising solutions.
To overcome this challenge, this paper proposes an efficient dynamic resource allocation framework for evolutionary bilevel optimization, named DRC-BLEA.
Compared to existing approaches, DRC-BLEA introduces a novel competitive quasi-parallel paradigm, in which multiple lower-level optimization tasks, derived from different upper-level individuals, compete for resources.
A continuously updated selection probability is used to prioritize execution opportunities to promising tasks.
Additionally, a cooperation mechanism is integrated within the competitive framework to further enhance efficiency and prevent premature convergence.
Experimental results compared with chosen state-of-the-art algorithms demonstrate the effectiveness of the proposed method.
Specifically, DRC-BLEA achieves competitive accuracy across diverse problem sets and real-world scenarios, while significantly reducing the number of function evaluations and overall running time.
\end{abstract}

\IEEEpeerreviewmaketitle

\section{Introduction}
\IEEEPARstart{B}{ilevel} optimization problems (BLOPs) are complicated due to  their hierarchical structure and the interactive nature of the decision-making process between the upper and lower levels, each with its own objective function involving both the upper- and lower-level decision variables.
Specifically, the upper-level decision variables $x_u$ serve as parameters for the lower-level problem, where only the corresponding lower-level optimal solution $x_l^*$ is considered a valid response.
Despite their complexity, bilevel problems frequently arise in real-world scenarios.
Examples include power grid configuration \cite{xia2021bilevel}, traffic network scheduling \cite{stoilova2022model}, resource pricing \cite{huang2021divide}, machine learning optimization \cite{rosales2022handling}, and many others \cite{yu2023engine}, \cite{louati2023embedding}, \cite{dempe2020bilevel}.
A notable application is hyper-parameter tuning in machine learning, which can be framed as a bilevel optimization problem.
At the upper level, the goal is to minimize validation loss by adjusting the hyper-parameters. The lower level, given the hyper-parameters, seeks to find a set of model parameters by minimizing training loss. In turn, the upper level evaluates and adjusts the hyper-parameters based on the performance of the learning model provided by the lower level. As problem scales increase, enhancing the efficiency of bilevel algorithms becomes increasingly essential.

Evolutionary algorithms (EAs) have been proven to be effective for solving bilevel optimization problems \cite{sinha2017review}.
Without making assumptions about problem characteristics, bilevel evolutionary algorithms (BLEAs) are considered more versatile than traditional mathematical approaches to handle various difficulties such as non-convexity and non-differentiable \cite{sinha2016solving}.
However, the application of evolutionary algorithms entails both advantages and challenges.
For each upper-level individual, the corresponding lower-level problem needs to be fully optimized to ensure the optimality.
Consequently, a large number of function evaluations (FEs) are required at the lower level, leading to high computational cost.
Assuming there are 100 upper-level individuals and each corresponding lower-level task evolves 100 individuals, even if only one iteration is performed for each lower-level task, a total of 10,000 FEs would be required.

To address these challenges, research on reducing the computational cost of evolutionary bilevel optimization has been active in recent years.
Meta-modeling methods were introduced to transform bilevel problems into single-level problems by constructing approximate mappings between the upper- and lower-level problems \cite{sinha2017evolutionary}.
However, these approximation methods are not always effective.
The lower-level reaction set mapping ($\psi$-mapping) is inaccessible when the lower-level optimal response to the upper-level decision is not single-valued \cite{sinha2014improved}, and the lower-level optimal value function mapping ($\varphi$-mapping) increases the dimensionality of variables and adds constraints to the transformed single-level problem.

In addition, some methods aim to reduce resource costs by exchanging information about optimal individuals.
This concept is common in co-evolutionary algorithms \cite{chaabani2018new}, where optimal individuals or coding information are shared between sub-populations or between upper- and lower-level populations \cite{chaabani2020co}. Recently, knowledge transfer has been implemented to facilitate the convergence of bilevel optimization \cite{chen2023evolutionary}.
For instance, marginal distributions are shared from the upper level to the lower level to enhance the quality of the lower-level initial population \cite{he2018evolutionary}.
In \cite{chen2021transfer}, \cite{gupta2015evolutionary}, and \cite{huang2023bilevel}, lower-level problems derived from multiple upper-level vectors are treated as similar tasks, where transfer occurs in the concurrent execution.

While the aforementioned methods have successfully accelerated the bilevel evolutionary process, they may incur massive resource expense when applied to problems with increased dimensions, which is inevitable in real-world applications \cite{liu2021investigating}, \cite{yimer2020proximal}.
The increase in variable dimensions expands the search space, requiring more evolutionary iterations, which leads to a significantly higher demand for function evaluations, especially at the lower level.
Most existing algorithms have been investigated on problems where both upper- and lower-level variables are fewer than 10, yet even under such conditions, obtaining quality results often requires over $1e^6$ FEs.
Therefore, improving the efficiency of BLEAs remains a significant challenge.

In essence, the inefficiency of BLEAs primarily stems from the large number of iterations executed in lower-level tasks. 
We can reconsider the rationality of complete lower-level optimization from the perspective of a typical evolution process, which inspires the motivation of this work.
A key operation in evolutionary computation is environmental selection, which is conducted at both levels.
For each upper-level individual $x_u$, the lower-level optimizer finds the best lower-level solution $x_l^*$ through iterative function evaluations and environmental selection. Subsequently, all the paired solutions ($x_u$, $x_l^*$) undergo upper-level function evaluation and selection.
Note that, even if all $x_u$ individuals are initially considered promising and their corresponding lower-level problems are fully optimized, some ($x_u$, $x_l^*$) pairs are not competitive.
As is common in practice, roughly half of these pairs are eliminated during upper-level selection.
In this context, the computational resources spent optimizing the lower-level tasks for these eliminated pairs are essentially wasted.
Current methods treat all lower-level problems derived from upper-level individuals equally, fully optimizing each one to find the lower-level optima.
As a result, although optimization of each lower-level task can be promoted, significant computational resources are wasted on those without potential.

Consequently, we consider the resource waste in the bilevel structure to be a significant obstacle to further improving optimization efficiency.
When handling a new upper-level population, it is more efficient not to execute all corresponding lower-level tasks with complete optimization equally.
Instead, the benefits of lower-level optimization can be maximized by prioritizing resource allocation to those lower-level tasks that are more likely to generate solutions capable of surviving the subsequent upper-level selection.

In this work, we propose a dynamic resource allocation framework for evolutionary bilevel optimization.
The lower-level optimization tasks derived from upper-level individuals are executed in a competitive paradigm, in which the execution priority of lower-level tasks is determined by a dynamically updated selection probability.
Generally, the contributions of this work can be summarized as follows:

$\bullet$ To eliminate the computational redundancy in BLEAs, we proposed a dynamic resource allocation framework with a competitive quasi-parallel paradigm for bilevel structure, which significantly reduces the resource consumption across problems of varying scales.

$\bullet$ To prioritize execution opportunities to tasks with higher competitiveness, we designed a probability-based task selection mechanism that combines the performance and evolving potential of each task, which enables the algorithm to improve efficiency while achieving competitive accuracy.

$\bullet$ To leverage the inherent similarity of competing tasks and enhance the overall competition quality, we implemented a knowledge transfer strategy based on the sharing of sampling parameters to promote cooperation in competition, which further improves the efficiency and prevents the algorithm from premature convergence.

The remainder of this paper is organized as follows:
Section II provides a brief introduction of the preliminaries and related work. 
Section III details the proposed method.
Section IV presents the experiments and analysis.
Finally, Section V concludes the paper with a discussion on future work.

\section{Preliminaries and Related Work}
\label{sec:Preliminaries-and-Related-Work}
\subsection{Bilevel Optimization}
A bilevel optimization problem can be formulated as: 
\begin{small}
\vspace{-0.6em}
\begin{equation}
\begin{gathered}
\underset{{x}_u \in X_u, {x}_l \in X_l}{\textbf{Min}} F\left({x}_u, {x}_l\right) \\
\textbf { s.t. } \quad {x}_l \in \textbf{argmin}\left\{f\left({x}_u, {x}_l\right): g_j\left({x}_u, {x}_l\right) \leq 0,\right. \\
j=1,2,\ldots, J\} \\
G_k\left({x}_u, {x}_l\right) \leq 0, k=1,2,\ldots, K,
\end{gathered}
\end{equation}
\end{small}where $F : \mathbb{R}^m \times \mathbb{R}^n \rightarrow \mathbb{R}$ is the upper-level objective function, and $f : \mathbb{R}^m \times \mathbb{R}^n \rightarrow \mathbb{R}$ is the lower-level objective function.
$G_k$ and $g_j$ are the upper- and lower-level constraints, respectively. 

The upper- and lower-level decision variables are taken into account in the objective functions and constraints at both levels. 
In addition to the explicit constraints, the lower-level problem acts as an implicit optimality constraint.
Specifically, only the optimal solution $x_l^*$ of $f(x_u,x_l)$ can be considered a valid candidate and paired with $x_u$ for upper-level evaluation.

\vspace{-0.5em}
\subsection{Related Work}
Early research on solving bilevel optimization problems mainly focus on classical mathematical methods.
Single-level reduction \cite{shi2005extended}, \cite{bard1990branch}, \cite{tuy1993global}, descent methods \cite{vicente1994descent}, \cite{savard1994steepest} and trust-region methods \cite{marcotte2001trust}, \cite{liu1998trust} have been widely used to address problems with favorable mathematical properties, such as linear, quadratic or convex functions.
Carefully crafted assumptions enable these algorithms to effectively solve problems with matching properties, but inevitably lead to performance deterioration when solving complex problems that fall outside these assumptions.
When the lower-level problem is convex and sufficiently regular, the Karush-Kuhn-Tucker (KKT) conditions can be used to simplify the BLOP into a single-level constrained optimization problem with Lagrangian and complementarity constraints.
The reformulated problem can then be addressed using penalty function methods \cite{lv2007penalty}, trust region methods \cite{colson2005trust}, or mixed-integer solvers \cite{edmunds1991algorithms}.
When dealing with linear or quadratic problems, KKT-reduction based methods perform well, but when the conditions are not applicable, it can lead to inaccuracy or infeasibility \cite{sinha2017review}.

Instead, evolutionary algorithms are capable of solving various problems in a black-box manner.
Research on BLEAs can be grouped into three main categories.

The first category is the single-level reduction method, which leverages the transformation techniques of classical mathematical approaches.
EAs are employed to solve the single-level problem transformed by methods such as the Karush-Kuhn-Tucker (KKT) conditions \cite{wan2013hybrid}, \cite{jiang2013application}, \cite{li2015genetic}.
However, the effectiveness of these methods is sensitive to the characteristics of the lower-level problem.

The second category is the approximation method.
Since each upper-level individual generates a lower-level problem, an intuitive way to simplify the hierarchical structure is to approximate the key relationship between upper and lower levels using various models including machine learning \cite{wang2023conditional}.
One approach is based on reaction set mapping \cite{sinha2014improved}, which approximates the optimal solution of the lower-level problem derived from the corresponding upper-level variables, converting the lower-level vector into a function of the upper-level variables \cite{angelo2014differential}.
Another approach is the optimal functional value mapping \cite{sinha2016solving}, which approximates the optimal function value of the corresponding lower-level problem, thereby transforming the lower-level problem into a constraint of the upper level.
Sinha \textit{et al.} \cite{sinha2020bilevel} discussed the pros and cons of the reaction set mapping and the optimal value mapping, and proposed BLEAQ-II, which adaptively selects one of the two approximate mappings according to the fitting quality.

The approximation methods perform well in some specific problems.
However, in real-world applications, algorithms often encounter black-box problems with varying types of function mappings.
Therefore, the general applicability of these methods is considered limited.
Additionally, concerns have been raised regarding the accuracy of approximation models, particularly when dealing with complex characteristics such as multi-modality.

The third category is based on the nested structure.
Without problem assumption, the nested methods have the best generalization ability.
Many typical EAs such as particle swarm optimization \cite{zhao2019nested} and differential evolution \cite{angelo2015study} are used at one or both levels.
However, even a simple bilevel problem can be computationally intensive for nested methods, so recent research in this category has focused on reducing computational cost and improving efficiency.

An intuitive way to improve computation efficiency is co-evolution \cite{cai2022cooperative}, \cite{chaabani2019transfer}.
For instance, Chaabani \textit{et al.} \cite{chaabani2018new} decomposed the population into a series of well-distributed sub-populations, and selected high-quality individuals for crossover to share information among them. 
In \cite{chaabani2020co}, co-evolution was performed at both levels.

Inspired by the strategy of solving expensive problems in single-level optimization, some methods incorporate surrogate models in bilevel optimization to approximate the original evaluation function or constraints \cite{angelo2019performance}. For example, Islam \textit{et al.} \cite{islam2017surrogate} used response surface models and Kriging to approximate the lower-level function. Kieffer \textit{et al.} \cite{kieffer2019tackling} generated a scoring function using a genetic programming hyper-heuristic.
Jiang \textit{et al.} \cite{jiang2023efficient} use Kriging models to approximate the objective function at the upper level and the optimal solution mapping at the lower level.
In \cite{gupta2016pareto}, both Pareto rank learning and fitness approximation were implemented at the upper level.
However, to some extent, surrogate models serve merely as alternatives to bypass the original evaluation, without reducing the actual number of environmental selections required in the evolutionary process. Furthermore, building and managing these models incur additional time and computational resources. The uncertainty in model performance may also impact the accuracy of the final results.

Exploring and transferring knowledge within the bilevel structure is an effective strategy to improve computational efficiency.
For example, He \textit{et al.} \cite{he2018evolutionary} proposed to share prior knowledge of the search distribution based on covariance matrix adaptation (CMA-ES) from the upper level to the lower level, providing a promising initial distribution for the lower level and significantly reducing the number of lower-level function evaluations.
In \cite{gupta2015evolutionary}, the lower-level problems corresponding to multiple upper-level individuals are treated as different tasks, and the evolutionary multitasking algorithm, known as MFEA \cite{gupta2015multifactorial}, is incorporated to handle multiple lower-level tasks simultaneously.
Recently, Chen \textit{et al.} \cite{chen2021transfer} optimized multiple lower-level problems concurrently through a parallel mechanism.
Knowledge transfer was conducted by selecting appropriate source and target domains, which promoted the convergence of the lower-level tasks and improved accuracy.

From the aforementioned work, it can be concluded that the nested evolutionary method is considered the most accurate and promising approach due to its insensitivity to problem characteristics.
However, its efficiency is limited by the large number of iterations required to achieve lower-level optimality.
To further improve the  computational efficiency, we propose a competitive bilevel framework in the next section, in which the computing resources are prioritized to the lower-level tasks that are most likely to generate promising $(x_u, x_l^*)$ pairs.

\section{Proposed Algorithm}
\label{sec:Proposed Algorithm}
In this section, we propose a dynamic resource allocation framework for evolutionary bilevel optimization, referred to as DRC-BLEA.
Overall, DRC-BLEA consists of three main components: 
1) dynamic resource competition, 2) selection probability update and 3) cooperation in competition.
Detailed descriptions of each component follow the overall framework description.

\begin{figure*}[htbp]
\setlength{\abovecaptionskip}{-0.1cm}
\centering
\includegraphics[width=0.9\textwidth]{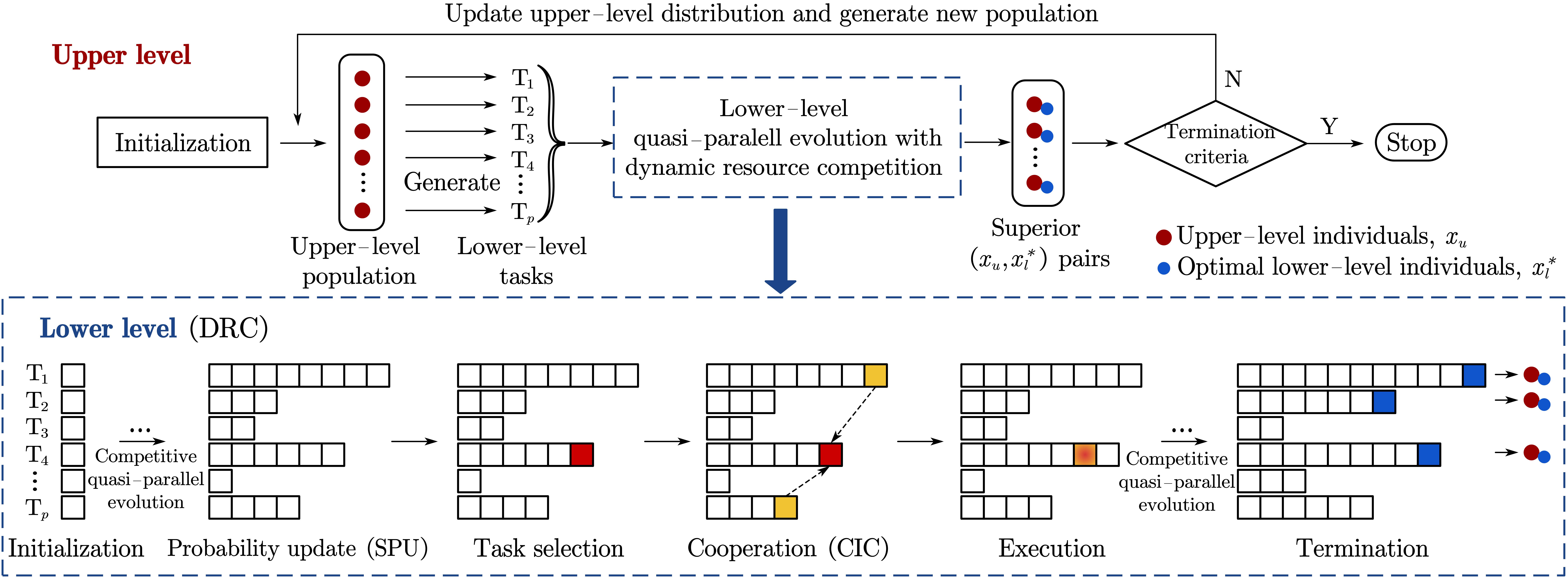}
\caption{Flowchart of the proposed DRC-BLEA. The squares represent the number of times each task has been executed at the lower level. The red square indicates that the corresponding task is selected to be executed, while the yellow square represents the cooperative tasks for the red one. The red-yellow mixed square indicates that the features of the selected task have been improved before execution through cooperation, and the blue square indicates that the task has met the lower-level termination condition.}
\label{fig:Flowchart}
\end{figure*}

\vspace{-0.5em}
\subsection{Overall Description}
The proposed framework is based on a nested bilevel evolutionary paradigm.
First, an upper-level population of size $p$ is initialized, which generates $p$ corresponding lower-level tasks.
Each lower-level task is then optimized using a lower-level population of size $q$.
Instead of solving the $p$ lower-level tasks sequentially, execution opportunities are dynamically assigned to each task through resource competition, allowing them to be gradually optimized in a quasi-parallel manner.
Subsequently, the upper-level population is updated with superior ($x_u$, $x_l^*$) pairs generated in the competitive quasi-parallel evolution.
The new population generates a new set of competing tasks, and such hierarchical iteration continues until the upper-level termination condition is met.
The overall flowchart of the proposed DRC-BLEA is presented in Fig.~\ref{fig:Flowchart}.

In the proposed framework, any population-based optimizer can be integrated to solve either the upper- or lower-level problem.
In this work, CMA-ES is adopted as the optimizer at both levels as a case study, and the algorithm is therefore referred to as DRC-CMA-ES.
CMA-ES is known for its efficiency and speed in solving optimization problems, which controls the evolution process with a covariance matrix.
The population is sampled from a multivariate Gaussian distribution generated by the covariance matrix, with the parameters of the Gaussian distribution adaptively adjusted via step size control and elite sample selection \cite{hansen2001completely}.
As suggested in \cite{he2018evolutionary}, both upper- and lower-level decision variables are considered in the upper-level evolution to facilitate the initialization of lower-level variables by sharing the marginal Gaussian distribution.
The framework of DRC-CMA-ES is illustrated in Algorithm ~\ref{alg:Framework of DRC-BLEA}.

When dealing with constrained problems, constraint violation (CV) is incorporated into the fitness evaluation.
Note that the upper-level constraint violation is calculated as the sum of the original upper- and lower-level constraint violation values, as a valid optimal solution must be feasible at both levels.
For environmental selection, the constraint Pareto preference is applied when comparing two candidate solutions \cite{deb2000efficient}, \cite{zhan2022survey}:
1) if $v_1$ is feasible and $v_2$ is infeasible, then $v_1$ is preferred.
2) if both $v_1$ and $v_2$ are feasible, then the one with a smaller objective function is preferred.
3) if both $v_1$ and $v_2$ are infeasible, then the one with a smaller constraint violation is preferred.

\vspace{-0.5em}
\subsection{Dynamic Resource Competition}
Instead of optimizing each lower-level task separately as in traditional nested evolution, the proposed DRC-CMA-ES optimizes multiple tasks in a competitive quasi-parallel manner.
The term ``quasi-parallel" referred here differs from the parallel framework described in \cite{chen2021transfer}, where each lower-level task evolves once in turn during each parallel round.
In DRC-CMA-ES, ``parallel" refers to the simultaneous handling of multiple lower-level tasks, including the concurrent activation and subsequent execution of multiple tasks.
At the beginning of each quasi-parallel evolution, $p$ lower-level tasks are optimized once in turn.
This ensures that all lower-level tasks are initiated and is also used to update the selection probability for subsequent executions.
However, the term ``quasi" indicates that after these $p$ tasks are activated, they are not executed with equal probability; instead, resources are dynamically allocated through competition.
Specifically, the amount of resources allocated corresponds to the number of function evaluations, embodied as the number of times the tasks being executed.

\begin{algorithm}[tbp]
\footnotesize
	\caption{Framework of DRC-CMA-ES}
     \label{alg:Framework of DRC-BLEA}
     \KwIn{the bilevel optimization problem.}
     \KwOut{the best solution.}
     Initialize the upper-level CMA-ES $cma_u$\;
     \While {the termination condition is not met}{
     Generate new upper-level population $Pop_u$ based on $cma_u$\;
     $Pop_e$ $=$ DRC($Pop_u$, $cma_u$)\;
     Update $cma_u$ with $Pop_e$\;
}
\Return the best solution.
\end{algorithm}

\begin{algorithm}[htbp]
\footnotesize
	\caption{Dynamic Resource Competition (DRC)}
     \label{alg:Dynamic Resource Competition}
     \KwIn{the upper-level population with $p$ individuals, $Pop_u$; the upper-level CMA-ES, $cma_u$.}
     \KwOut{the elite population, $Pop_e$;}
     $T_{count}$ = 0, $Arc$ = $\emptyset$, $Pop_e$ = $\emptyset$\;
     \For{i = \rm 1 to $p$}{
     Initialize the lower-level CMA-ES parameter $cma_l^i$ with the marginal Gaussian distribution of $cma_u$ for upper-level individual $x_u^i$ \;
     Generate the lower-level population $Pop_l^i$ based on $cma_l^i$\ and evaluate the individuals with lower-level FEs\;
     Update $cma_l^i$ with the elite individuals\;
     Pick out the best lower-level individual $x_l^{i,1}$ in $Pop_l^i$\;
     Evaluate ($x_u^i,x_l^{i,1}$) with upper-level FEs and store it into archive $Arc$;}
     Initialize the selection probability set as $P$ = SPU($Arc$, $p$)\;
     \While {$T_{count}$ \textless $\lfloor p/2 \rfloor$}{
     \For{j = \rm 1 to $p$}{
     Select a task $T_m$ by roulette wheel based on the selection probability set $P$\;
     \eIf{$T_m$ has been executed for three or more times}{
     ($cma_l^m$, $x_l^{navi}$) = CIC($T_m$, $Arc$, $cma_l$)\;
     Generate the population $Pop_l^m$ based on $cma_l^m$ and $x_l^{navi}$\;}
     {Generate the population $Pop_l^m$ based on $cma_l^m$\;}
     Evaluate the population $Pop_l^m$ with lower-level FEs, and update $cma_l^m$\;
     \If{the lower-level optimum $x_l^{m,t}$ of $T_m$ is updated}{
     Evaluate newly generated ($x_u^m,x_l^{m,t}$) with upper-level FEs\;
     }
     Store ($x_u^m,x_l^{m,t}$) into archive $Arc$\;
     \If{the lower-level termination condition of $T_m$ is met}{
     $x_l^{m*}$ = $x_l^{m,t}$, $Pop_e$ = $Pop_e$ $\cup$ ($x_u^m,x_l^{m*}$) \;
     Delete $T_m$ from the calculation of SPU\;
     $T_{count} = T_{count}+ 1$\;
     Update the selection probability set as $P$ = SPU($Arc$, $p-T_{count}$)\;}}
     Update the selection probability set as $P$ = SPU($Arc$, $p-T_{count}$)\;}
\Return ($Pop_e$)\;
\end{algorithm}

The schematic diagram of the dynamic resource competition mechanism is shown in the dashed box of Fig.~\ref{fig:Flowchart}. 
Each round of the competitive evolution performs $p$ executions, where $p$ represents the number of competing tasks and corresponds to the size of upper-level population.
The task selected in each execution is determined by a roulette wheel based on the selection probabilities of all competing tasks.
After each round of $p$ selections and executions, the selection probabilities are reevaluated according to the performance of each task.

Due to varying characteristics and differing amounts of  resources allocated, the tasks achieve lower-level convergence at different time instants.
When a task meets the lower-level termination condition, it is withdrawn from the competition.
The selection probabilities of the remaining tasks are then reevaluated, and the completed tasks are no longer assigned selection probabilities.

In a typical evolution process, commonly half of the individuals are retained in the selection operation. Naturally, in DRC-CMA-ES, once half of the tasks have met the termination condition, this set of competitive quasi-parallel evolution will be terminated.
In a way, the selection operation is implicitly embedded within the competition process.
The half of the tasks that converge earlier, due to better performance with more execution resources allocated, are considered the winners, and the resulting ($x_u, x_l^*$) pairs are directly used to update the population for the next upper-level iteration. 
The pseudo-code of the dynamic resource competition is presented in Algorithm ~\ref{alg:Dynamic Resource Competition}.

\vspace{-0.5em}
\subsection{Selection Probability Update}
In this section, we describe how the selection probability is updated based on the competitiveness of each task, allowing more computing resources to be allocated to tasks that are more likely to generate superior ($x_u, x_l^*$) pairs for the upper level.
Since the ultimate intention of competition among tasks is to survive the upper-level selection, task competitiveness is evaluated based on upper-level fitness rather than lower-level fitness.

The task competitiveness is evaluated from two aspects. 
The first aspect is about the upper-level fitness of the ($x_u,x_l^t$) paired by the upper-level variables and the temporary optimal lower-level variables ($t$ denotes temporary) obtained in the tasks.
A relatively high fitness of ($x_u,x_l^t$) pairs may result from preferable upper-level variables or lower-level variables, or even superior at both levels.
Such pairs are more likely to prevail in the competition driven by upper-level fitness.

The second aspect is regarding the evolving potential of the tasks.
In the competitive quasi-parallel evolution, only the lower-level variables of selected tasks are optimized. 
Successful evolution results in improved lower-level variables $x_l^t$, which correspondingly alters the upper-level fitness of the ($x_u, x_l^t$) pair.
If the change is significant, it indicates that there remains room for improvement at the lower level of the task.
Therefore, changes in the upper-level fitness reveal the evolving potential of tasks.
With high evolving potential, even if the temporary fitness of a task is relatively low, given more resource, it is likely to come from behind in the following competition.

Therefore, the selection probability consists of three parts: basic probability, performance probability, and potential probability.
The basic probability ensures that all tasks have a baseline chance of being selected, regardless of their performance at any point in the competition process.
This prevents potential tasks that are currently less competitive from being ignored. The basic probability for each task is given as:

\begin{small}
\vspace{-0.6em}
\begin{equation}
\label{equation:Pkbs}
P^k_{bs}=1 / K, k=1,2, \ldots, K
\end{equation}
\end{small}where $K$ represents the number of current competing tasks, and $K$ is less than or equal to the initial number of tasks $p$ at the start of competitive quasi-parallel evolution.

The performance probability is evaluated by the upper-level fitness of the temporary optimal pair.
The higher the upper-level fitness of the ($x_u^k,x_l^{k,t}$) pair obtained by task $k$, the higher the performance probability will be assigned.
Since the number of executions varies across tasks, we first calculate a competing fitness for each task using a weighted sum of its historical fitness. Inspired by DEORA \cite{li2022evolutionary}, the competing fitness is defined as:

\begin{small}
\vspace{-0.6em}
\begin{equation}
\label{equation:CFk}
CF^k=\frac{1}{\sum_{t=1}^{T^k} \gamma^{T^k-t}} \sum_{t=1}^{T^k} \gamma^{T^k-t} F(x_u^k,x_l^{k,t})
\end{equation}
\end{small}where $T^k$ represents the number of times task $k$ has been selected and executed. 0 $\leq \gamma \leq$ 1 is the discount factor, and it leads to more attention paid to the most recent fitness. 
In the following experimental study,  $\gamma$ is set to 0.5.
The scale of competing fitness depends not only on the problem and the optimizer, but also on the evolution stage.
Therefore, each time we update the performance probability, we set the worst competing fitness of all the current tasks as the baseline, and then assign the probabilities to other tasks based on their fitness advantage over the baseline $CF^{min}$.
The performance probability $P_{pf}^k$ of task $k$ is updated as:

\begin{small}
\vspace{-0.6em}
\begin{equation}
\label{equation:Pkpf}
P_{p f}^k=\frac{CF^k- CF^{min} }{\sum_{k=1}^K CF^k- CF^{min}}, k=1,2, \ldots, K
\end{equation}
\end{small}

\begin{algorithm}[tbp]
\footnotesize
	\caption{Selection Probability Update (SPU)}
     \label{alg:Selection Probability Update}
     \KwIn{the archive that stores the historical optimal ($x_u^k,x_l^{k,t}$) pairs and the corresponding upper-level fitness for each task, $Arc$; the number of current competing tasks, $K$}
     \KwOut{the selection probability set, $P$;}
     \For{k = \rm 1 to $K$}{
     Calculate the competing fitness $CF^k$ by (\ref{equation:CFk}) and the competing potential $CP^k$ by (\ref{equation:CPk}) based on the archive $Arc$\;
     }
     \For{k = \rm 1 to $K$}{
     Assign basic probability $P_{bs}^k$ for task $k$ by (\ref{equation:Pkbs})\;
     Calculate the performance probability $P_{pf}^k$ by (\ref{equation:Pkpf}) and the potential probability $P_{pt}^k$ by (\ref{equation:Pkpt})\;
     Update the selection probability $P^k$ for task $k$ by (\ref{equation:Pk})\;
     }
\Return $P = \{P^1,..., P^K\}$\;
\end{algorithm}

The potential probability focuses on the evolving potential revealed by the fitness variation of the tasks.
It is worth noting that a successful execution in the parallel competitive evolution will improve the lower-level fitness $f$($x_u^k,x_l^{k,t}$) of the selected task $k$.
However, the upper-level fitness $F$($x_u^k,x_l^{k,t}$) may either improve or deteriorate.
This variability arises from the fact that there may be conflict or cooperation between the upper- and lower-level objective functions, which is determined by the characteristics of the problem itself and cannot be known in advance.
Therefore, we cannot assume that the upper-level fitness changes in a specific direction.

To capture the evolving potential, we first calculate the variation rate of the upper-level fitness before and after each execution.
From a global perspective of the entire competitive evolution set, we further define the potential reward and potential penalty.
If the fitness of the new pair ($x_u^k,x_l^{k,t}$) generated by executing the selected task is better than the global optimal fitness of all parallel tasks before this execution, the improvement rate of $F$($x_u^k,x_l^{k,t}$) to the previous temporary global optimal fitness $F_{gb}^{t-1}$ is counted as a reward.
Conversely, if $F$($x_u^k,x_l^{k,t}$) is worse than the global worst fitness $F_{gw}^{t-1}$, the deterioration rate is counted as a penalty.
Consequently, the evolving potential is defined as:

\begin{small}
\vspace{-0.6em}
\begin{equation}
\label{equation:Ptk}
\begin{aligned}
PT^k=& \frac{F\left(x_u^k, x_l^{k,t}\right)-F\left(x_u^k, x_l^{k,t-1}\right)}{\left|F\left(x_u^k, x_l^{k,t-1}\right)\right|}\\
&+\max \left\{\frac{F\left(x_u^k, x_l^{k,t}\right)-F_{gb}^{t-1}}{\left|F_{gb}^{t-1}\right|}, 0\right\} \\
&+\min \left\{\frac{F\left(x_u^k, x_l^{k,t}\right)-F_{gw}^{t-1}}{\left|F_{gw}^{t-1}\right|}, 0\right\}
\end{aligned}
\end{equation}
\end{small}

Similarly, the competing potential is defined as:
\begin{small}
\vspace{-0.6em}
\begin{equation}
\label{equation:CPk}
CP^k=\frac{1}{\sum_{t=2}^{T^k} \gamma^{T^k-t}} \sum_{t=2}^{T^k} \gamma^{T^k-t} PT^k
\end{equation}
\end{small}

As mentioned above, the direction of competing potential can be either positive or negative.
But in general, the greater the tendency of fitness to improve or the smaller the tendency of fitness to deteriorate, the more likely the task is to become competitive in subsequent parallel evolution.
Therefore, we assign higher probability to tasks with greater potential in the positive direction.
We use an $\epsilon$-based exponential function ($\epsilon$ = 1.1) to transform the scale of the competing potential to range (0,+$\infty$).
The potential probability of each task is then calculated by the ratio of its exponential competing potential to the total exponential competing potential of all tasks as:

\begin{small}
\vspace{-0.6em}
\begin{equation}
\label{equation:Pkpt}
P_{pt}^k=\frac{\epsilon^{CP^k}}{\sum_{k=1}^K \epsilon^{CP^k}}, k=1,2, \ldots, K
\end{equation}
\end{small}

For each task, the probability of being selected is a weighted sum of the three probabilities described above:

\begin{small}
\vspace{-1em}
\begin{equation}
\label{equation:Pk}
P^k=w_{bs} \cdot P_{bs}^k+w_{pf} 
\cdot P_{pf}^k+w_{pt} \cdot P_{pt}^k, k=1,2, \ldots, K
\end{equation}
\end{small}

Among the three probabilities, the basic probability $P_{bs}^k$ only needs to be assigned a small weight, $w_{bs}$, to provide a minimum probability assurance for each task.
The weight for the performance probability $P_{pf}^k$, $w_{pf}$, should be set relatively high as it directly reveals the current competitiveness of the task.
The potential probability $P_{pt}^k$ facilitates the exploration of competitive evolution mechanism.
Accordingly, the weights in DRC-CMA-ES are set as $w_{bs}$= 0.1, $w_{pf}$= 0.7 and $w_{pt}$= 0.2. 
The rationale behind these weight settings is verified and discussed in the supplementary material.
The pseudo-code of the selection probability update (SPU) is provided in Algorithm ~\ref{alg:Selection Probability Update}.

\begin{algorithm}[tbp]
\footnotesize
	\caption{Cooperation in Competition (CIC)}
     \label{alg:Cooperation in Competition}
     \KwIn{The selected task as the target, $T_m$;
     the archive that stores the historical optimal ($x_u^k,x_l^{k,t}$) pairs and the corresponding upper-level fitness for each task, $Arc$;
     the set of lower-level CMA-ES parameters for all competing tasks, $cma_l$;}
     \KwOut{Updated lower-level CMA-ES parameters of task $m$, $cma_l^m$;
     a lower-level navigational solution for task $m$, $x_l^{navi}$;}
     Select qualified source tasks $T_n$ for the target task $T_m$\;
     \eIf{no tasks meet the criteria of being a source task}{
     \Return $cma_l^m$ without cooperation\;}
     {Calculate the cooperation intensity $CI(n)$ of each source task $T_n$ by (\ref{equation:CIS})\;
     Update the lower-level CMA-ES parameters of task $T_m$
     by (\ref{equation:mtct})\;
     Select the lower-level optimal solution of the task with the highest cooperation intensity as $x_l^{navi}$\;
\Return $cma_l^m$, $x_l^{navi}$\;}
\end{algorithm}

\vspace{-0.5em}
\subsection{Cooperation in Competition}
In the competitive quasi-parallel framework described above, the competing lower-level tasks corresponding to different upper-level individuals originate form the same bilevel problem, making them inherently a class of problems sharing the same structure.
The primary difference between these tasks is that some parameters in the lower-level objective or constraint functions depend on the respective upper-level individuals.

With similar upper-level variables as  parameters, the lower-level problems may share similar landscapes and similar optimal solutions.
Moreover, the competing tasks are always in different convergence stages due to variation in their initial lower-level variables and the resources allocated.
Such concurrent similarity and difference between tasks inspire cooperation in competition, which can enhance the quality of the competitors and thereby facilitate overall convergence.
The cooperation is performed by transferring knowledge from similar tasks in more advanced convergence stage to those in earlier stages, allowing tasks to move faster toward the optima or escape from the local optima.

To perform beneficial cooperation and prevent negative transfer, it is crucial to measure the similarity between tasks and the differences in their convergence stages.
Since the upper-level individuals impact some of the parameters in the corresponding lower-level tasks, the similarity between tasks is compared based on the distance between their upper-level individuals.
The distance $d_u$ is measured by $Lp$-norm with $p$ = 1/$m$ for diversity maintenance as recommended in \cite{song2021kriging}.
The difference in convergence stages between the lower-level tasks is independent of the absolute upper- or lower-level fitness, as their optimal value may be different.
Considering the characteristics of the heuristic algorithm during the optimization process, where changes occur rapidly in the earlier stages and slowly in the later stages \cite{wang2017population}.
In DRC-CMA-ES, the convergence stages of lower-level tasks are compared based on the fluctuations in their sampling features. 
The smaller the fluctuations in the CMA-ES mean vector over successive executions, the more likely a task is to approach lower-level convergence.
To capture the latest fluctuations, the standard deviation of the lower-level mean vector over the last three executions is used for comparison.

In the competitive quasi-parallel framework, cooperation between tasks is unidirectional.
When a task is selected for execution, it is considered as the target task.
The source tasks for cooperation are selected based on distance and the standard deviation of the CMA-ES mean vector.
A smaller distance indicates higher similarity, while a smaller standard deviation reflects better convergence.
Tasks with a smaller standard deviation in their CMA-ES mean vector than the target task are selected first as candidates for source tasks.
These candidates are then further filtered based on the distances between their upper-level variables and that of the target task, with only those in the closest half of all tasks being considered.
Additionally, to preserve diversity, only tasks that have been executed for three or more times can be considered as either target or source tasks.

When multiple source tasks are available, the target task cooperates with each source task at different intensities. The cooperation intensity depends on both the standard deviation and the distance.
Superior source tasks, which exhibit smaller fluctuations and higher similarity, are assigned higher cooperation intensities.
The cooperation intensity is defined as:

\begin{small}
\vspace{-1em}
\begin{equation}
\label{equation:CIS}
CI(s)=1-\alpha \frac{s t d(s)}{s t d(t) +\sum_{s=1}^S s t d(s)}-(1-\alpha) \frac{d_u(s)}{\sum_{s=1}^S d_u(s)}
\end{equation}
\end{small}where $S$ is the number of selected source tasks, and $std(s)$ represents the standard deviation of CMA-ES mean vector of task $s$. $d_{u}(s)$ denotes the distance between the upper-level individuals of  task $s$ and the target task $t$.
The coefficient 0 $\leq \alpha \leq$ 1 balances the weight between upper-level similarity and convergence phase. 
As the setting of $\alpha$ is problem specific, we set it as 0.5 to enhance versatility.

In DRC-CMA-ES, key sampling features, specifically the mean vector and the covariance matrix, are shared between cooperative tasks.
Based on the cooperation intensity, the mean vector and covariance matrix of the target CMA-ES are updated through a weighted mixture:

\begin{small}
\vspace{-1em}
\begin{equation}
\label{equation:mtct}
\begin{gathered}
m_t=\left(1-\alpha \frac{s t d(t)}{s t d(t)+\sum_{s=1}^S s t d(s)}\right) \cdot m_t+\sum_{s=1}^S C I(s) \cdot m_s \\
C_t=\left(1-\alpha \frac{s t d(t)}{s t d(t)+\sum_{s=1}^S s t d(s)}\right) \cdot C_t+\sum_{s=1}^S C I(s) \cdot C_s
\end{gathered}
\end{equation}
\end{small}

The cooperation based on the sampling features expedites the convergence of the target task by steering the sampling area closer to the optima.
However, due to the inherent randomness of sampling, even with superior sampling features, promising variables may not be sampled with certainty.
To further enhance the cooperation, the lower-level optimal solution of the task with the highest cooperation intensity is selected as a navigational solution $x_l^{navi}$.
Following the sampling of new lower-level variables for the target task based on the cooperatively updated sampling features, $x_l^{navi}$ is taken into account the fitness ranking and the sampling features update of the target task. 
The navigational solution works as a promising solution that has been explicitly sampled to improve the quality of the next sampling for the target task.
To prevent the tasks from losing diversity, $x_l^{navi}$ will not be adopted as the new optimum of the target task, even if it outperforms other sampled solutions.
The pseudo-code of cooperation in competition is provided in Algorithm ~\ref{alg:Cooperation in Competition}.

In the cooperation mechanism, knowledge sharing can be conducted from both directions between tasks with different competitiveness, rather than exclusively from high-competitiveness tasks to low-competitiveness ones.
Thus, the cooperation will not lead to homogenization of task results.
Instead, the multi-directional sharing facilitates lower-level convergence and improves the overall competition quality, thereby helping the algorithm accomplish the competitive quasi-parallel procedure with less resources and reducing the risk of premature convergence.

\section{Experimental Study}
\subsection{Benchmark Problems and Performance Indicators}
The algorithms are evaluated on the widely recognized SMD test suite \cite{sinha2014test} with 12 scalable bilevel problems, and the TP test suite \cite{sinha2017evolutionary} with 10 non-scalable problems.
SMD introduces challenges in terms of controlled convergence, interaction between the upper- and lower-level optimization tasks, multiple global solutions and constraints within the problems.
To assess the scalability of the algorithms, three different dimension settings are applied to each problem by varying the combinations of $m$ (upper-level variables) and $n$ (lower-level variables): ($m$ = 2, $n$ = 3), ($m$ = 10, $n$ = 10), and ($m$ = 30, $n$ = 30).
Most of the objective and constraint functions in TP are linear or quadratic, and the variable dimensions are fixed.
The true optimal function values ($F^*$,  $f^*$) for the problems are independent of the variable dimensions, and they are known for performance comparison.

In the numerical experiments, accuracy and the number of function evaluations (FEs) are used as the performance indicators:

$\bullet$ The accuracy of an obtained function value is defined as its absolute distance from the optimal function value ($F^*$,  $f^*$).
Accuracy is calculated at both levels, i.e.,
$Acc_u$ = $\left|F-F^*\right|$ for the upper level and $Acc_l$ = $\left|f-f^*\right|$ for the lower level.
In the comparison study, the best results ($F^{best}$,  $f^{best}$) obtained at the termination are used for accuracy calculation, and the results with accuracy less than $1e^{-6}$ are recorded as $1e^{-6}$ to filter out subtle differences.

$\bullet$ The total number of FEs at the upper and lower levels are denoted as $FEs_u$ and $FEs_l$, respectively.
Due to different strategies, the proportions of $FEs_u$ and $FEs_l$ consumed by the algorithms at termination is different. 
For a fair comparison, the overall number of function evaluations is considered, i.e., $FEs = FEs_u + FEs_l$.

\begin{table}[b]
\vspace{-1em}
\setlength{\abovecaptionskip}{-1pt}
\scriptsize
  \centering
  \caption{Termination conditions for upper and lower levels on different problems}
   \label{table:termination criteria}
    \renewcommand\arraystretch{0.5}
    \begin{tabular}{ccccc}
    \toprule
    Problem (scale) & $FEs_u^{max}$ & $FEs_u^{var}$ & $FEs_l^{max}$ & $FEs_l^{var}$ \\
    \midrule
    SMD ($m$ = 2, $n$ = 3) & 2500  & 350   & 250   & 25 \\
    SMD ($m$ = 10, $n$ = 10) & 5000  & 750   & 500   & 50 \\
    SMD ($m$ = 30, $n$ = 30) & 12500 & 750   & 1000  & 50 \\
    TP ($m$ and $n$ are fixed) & 5000 & 350 & 250 & 35 \\
    \bottomrule
    \end{tabular}%
   \vspace{-1em}
\end{table}%

\begin{table*}[htbp]
  \centering
  \caption{Median and IQR results of the Accuracy $Acc_u$ and $Acc_l$ achieved by the proposed DRC-CMA-ES and comparison algorithms in 21 runs on SMD problems with dimension  ($m$ = 2, $n$ = 3). The best result in each case is highlighted with a gray background.}
    \label{table:Acc_m2n3}
     \resizebox{\textwidth}{!}{
\begin{tabular}{cccccccccc}
\toprule
    \multicolumn{2}{c}{($m$ = 2, $n$ = 3)} & DRC-CMA-ES (ours) & BL-CMA-ES & TLEA-CMA-ES & NBLEA & BLEAQ-II & GO-CMA-ES & MFBLA & SA-BL-IMODE \\
    \midrule
    \multirow{2}[2]{*}{SMD1} & $Acc_u$  & \cellcolor[rgb]{.851,.851,.851}1.00E-06(0.00E+00) & \cellcolor[rgb]{.851,.851,.851}1.00E-06(0.00E+00)$\approx$ & \cellcolor[rgb]{.851,.851,.851}1.00E-06(0.00E+00)$\approx$ & 3.47E-05(5.09E-05)+ & \cellcolor[rgb]{.851,.851,.851}1.00E-06(0.00E+00)$\approx$ & \cellcolor[rgb]{.851,.851,.851}1.00E-06(0.00E+00)$\approx$ & 3.81E-06(2.86E-05)+ & \cellcolor[rgb]{.851,.851,.851}1.00E-06(0.00E+00)$\approx$ \\
          & $Acc_l$  & \cellcolor[rgb]{.851,.851,.851}1.00E-06(0.00E+00) & \cellcolor[rgb]{.851,.851,.851}1.00E-06(0.00E+00)$\approx$ & \cellcolor[rgb]{.851,.851,.851}1.00E-06(0.00E+00)$\approx$ & 1.13E-05(2.30E-05)+ & \cellcolor[rgb]{.851,.851,.851}1.00E-06(0.00E+00)$\approx$ & \cellcolor[rgb]{.851,.851,.851}1.00E-06(0.00E+00)$\approx$ & 2.14E-06(2.12E-06)+ & \cellcolor[rgb]{.851,.851,.851}1.00E-06(0.00E+00)$\approx$ \\
    \midrule
    \multirow{2}[2]{*}{SMD2} & $Acc_u$  & \cellcolor[rgb]{.851,.851,.851}1.00E-06(0.00E+00) & \cellcolor[rgb]{.851,.851,.851}1.00E-06(0.00E+00)$\approx$ & \cellcolor[rgb]{.851,.851,.851}1.00E-06(0.00E+00)$\approx$ & 4.10E-05(7.16E-05)+ & 2.32E-05(2.95E-05)+ & \cellcolor[rgb]{.851,.851,.851}1.00E-06(0.00E+00)$\approx$ & 5.64E-06(9.78E-06)+ & \cellcolor[rgb]{.851,.851,.851}1.00E-06(0.00E+00)$\approx$ \\
          & $Acc_l$  & \cellcolor[rgb]{.851,.851,.851}1.00E-06(2.07E-06) & 2.04E-06(7.69E-06)$\approx$ & 1.07E-06(2.31E-06)$\approx$ & 9.20E-06(3.36E-05)+ & 3.09E-05(4.67E-05)+ & \cellcolor[rgb]{.851,.851,.851}1.00E-06(0.00E+00)$\approx$ & 1.69E-06(4.58E-06)+ & \cellcolor[rgb]{.851,.851,.851}1.00E-06(0.00E+00)$\approx$ \\
    \midrule
    \multirow{2}[2]{*}{SMD3} & $Acc_u$  & \cellcolor[rgb]{.851,.851,.851}1.00E-06(0.00E+00) & \cellcolor[rgb]{.851,.851,.851}1.00E-06(0.00E+00)$\approx$ & \cellcolor[rgb]{.851,.851,.851}1.00E-06(0.00E+00)$\approx$ & 7.68E-05(1.07E-04)+ & \cellcolor[rgb]{.851,.851,.851}1.00E-06(0.00E+00)$\approx$ & \cellcolor[rgb]{.851,.851,.851}1.00E-06(0.00E+00)$\approx$ & 2.27E-05(2.56E-05)+ & \cellcolor[rgb]{.851,.851,.851}1.00E-06(0.00E+00)$\approx$ \\
          & $Acc_l$  & \cellcolor[rgb]{.851,.851,.851}1.00E-06(3.73E-08) & \cellcolor[rgb]{.851,.851,.851}1.00E-06(1.05E-06)$\approx$ & \cellcolor[rgb]{.851,.851,.851}1.00E-06(3.89E-07)$\approx$ & 3.07E-05(7.12E-05)+ & \cellcolor[rgb]{.851,.851,.851}1.00E-06(5.84E-07)$\approx$ & \cellcolor[rgb]{.851,.851,.851}1.00E-06(0.00E+00)$\approx$ & 2.09E-06(4.39E-06)+ & \cellcolor[rgb]{.851,.851,.851}1.00E-06(0.00E+00)$\approx$ \\
    \midrule
    \multirow{2}[2]{*}{SMD4} & $Acc_u$  & \cellcolor[rgb]{.851,.851,.851}1.00E-06(0.00E+00) & \cellcolor[rgb]{.851,.851,.851}1.00E-06(0.00E+00)$\approx$ & \cellcolor[rgb]{.851,.851,.851}1.00E-06(0.00E+00)$\approx$ & 2.05E-05(5.26E-05)+ & \cellcolor[rgb]{.851,.851,.851}1.00E-06(2.40E-06)+ & \cellcolor[rgb]{.851,.851,.851}1.00E-06(0.00E+00)$\approx$ & 2.84E-06(3.16E-05)+ & \cellcolor[rgb]{.851,.851,.851}1.00E-06(0.00E+00)$\approx$ \\
          & $Acc_l$  & 2.70E-06(4.01E-06) & 3.97E-06(5.79E-06)+ & 4.16E-06(9.42E-06)+ & 1.33E-05(2.51E-05)+ & 1.21E-05(6.77E-05)+ & 3.22E-06(3.15E-06)+ & \cellcolor[rgb]{.851,.851,.851}{1.00E-06(2.94E-05)-} & \cellcolor[rgb]{.851,.851,.851}{1.00E-06(0.00E+00)-} \\
    \midrule
    \multirow{2}[2]{*}{SMD5} & $Acc_u$  & \cellcolor[rgb]{.851,.851,.851}1.00E-06(0.00E+00) & \cellcolor[rgb]{.851,.851,.851}1.00E-06(0.00E+00)$\approx$ & \cellcolor[rgb]{.851,.851,.851}1.00E-06(0.00E+00)$\approx$ & 1.30E-05(5.72E-05)+ & 1.51E-05(6.75E-05)+ & 3.52E-02(6.36E-02)+ & 9.26E-06(1.52E-05)+ & \cellcolor[rgb]{.851,.851,.851}1.00E-06(0.00E+00)$\approx$ \\
          & $Acc_l$  & \cellcolor[rgb]{.851,.851,.851}1.14E-06(1.50E-06) & 1.23E-06(8.27E-07)$\approx$ & 1.87E-06(4.69E-06)+ & 8.90E-06(3.98E-05)+ & 1.93E-05(6.83E-05)+ & 3.52E-02(6.36E-02)+ & 2.14E-06(5.40E-06)+ & 1.22E-06(6.27E-07)$\approx$ \\
    \midrule
    \multirow{2}[2]{*}{SMD6} & $Acc_u$  & \cellcolor[rgb]{.851,.851,.851}1.00E-06(0.00E+00) & \cellcolor[rgb]{.851,.851,.851}1.00E-06(0.00E+00)$\approx$ & \cellcolor[rgb]{.851,.851,.851}1.00E-06(0.00E+00)$\approx$ & 9.19E-04(8.54E-04)+ & \cellcolor[rgb]{.851,.851,.851}1.00E-06(0.00E+00)$\approx$ & \cellcolor[rgb]{.851,.851,.851}1.00E-06(0.00E+00)$\approx$ & 8.02E+00(8.09E+00)+ & \cellcolor[rgb]{.851,.851,.851}1.00E-06(0.00E+00)$\approx$ \\
          & $Acc_l$  & \cellcolor[rgb]{.851,.851,.851}1.00E-06(0.00E+00) & \cellcolor[rgb]{.851,.851,.851}1.00E-06(0.00E+00)$\approx$ & \cellcolor[rgb]{.851,.851,.851}1.00E-06(0.00E+00)$\approx$ & 7.91E-05(3.62E-04)+ & \cellcolor[rgb]{.851,.851,.851}1.00E-06(0.00E+00)$\approx$ & \cellcolor[rgb]{.851,.851,.851}1.00E-06(0.00E+00)$\approx$ & 2.02E+00(4.45E+00)+ & \cellcolor[rgb]{.851,.851,.851}1.00E-06(0.00E+00)$\approx$ \\
    \midrule
    \multirow{2}[2]{*}{SMD7} & $Acc_u$  & \cellcolor[rgb]{.851,.851,.851}1.00E-06(9.82E-02) & 9.17E-02(9.82E-02)+ & 9.82E-02(9.82E-02)+ & 1.29E-05(9.82E-02)$\approx$ & 3.76E-06(9.82E-02)$\approx$ & \cellcolor[rgb]{.851,.851,.851}1.00E-06(0.00E+00)$\approx$ & 1.10E-05(9.82E-02)+ & \cellcolor[rgb]{.851,.851,.851}1.00E-06(0.00E+00)$\approx$ \\
          & $Acc_l$  & 2.33E-06(2.44E+02) & 1.22E+02(2.44E+02)+ & 2.44E+02(2.44E+02)+ & 1.48E-06(2.44E+02)$\approx$ & 5.96E-06(2.44E+02)$\approx$ & \cellcolor[rgb]{.851,.851,.851}1.00E-06(0.00E+00)- & \cellcolor[rgb]{.851,.851,.851}1.00E-06(2.45E+02)- & \cellcolor[rgb]{.851,.851,.851}1.00E-06(0.00E+00)- \\
    \midrule
    \multirow{2}[2]{*}{SMD8} & $Acc_u$  & 1.54E-05(5.83E-05) & \cellcolor[rgb]{.851,.851,.851}1.00E-06(3.15E-06)- & \cellcolor[rgb]{.851,.851,.851}1.00E-06(1.39E-06)- & 1.57E-03(2.39E-03)+ & 1.23E-05(6.79E-04)$\approx$ & 2.45E-01(1.86E+00)+ & 4.38E-03(5.45E-03)+ & \cellcolor[rgb]{.851,.851,.851}1.00E-06(0.00E+00)- \\
          & $Acc_l$  & \cellcolor[rgb]{.851,.851,.851}1.00E-06(1.20E-06) & \cellcolor[rgb]{.851,.851,.851}1.00E-06(0.00E+00)$\approx$ & \cellcolor[rgb]{.851,.851,.851}1.00E-06(0.00E+00)$\approx$ & 1.39E-04(4.02E-04)+ & 1.66E-05(1.70E-04)$\approx$ & 2.45E-01(1.86E+00)+ & 6.20E-03(1.20E-02)+ & 1.48E-03(1.11E-02)+ \\
    \midrule
    \multirow{2}[2]{*}{SMD9} & $Acc_u$  & \cellcolor[rgb]{.851,.851,.851}1.00E-06(0.00E+00) & \cellcolor[rgb]{.851,.851,.851}1.00E-06(0.00E+00)$\approx$ & \cellcolor[rgb]{.851,.851,.851}1.00E-06(0.00E+00)$\approx$ & 7.64E-03(3.56E-01)+ & 3.17E-05(5.12E-05)+ & \cellcolor[rgb]{.851,.851,.851}1.00E-06(0.00E+00)$\approx$ & 5.19E-06(1.60E-05)+ & \cellcolor[rgb]{.851,.851,.851}1.00E-06(0.00E+00)$\approx$ \\
          & $Acc_l$  & \cellcolor[rgb]{.851,.851,.851}1.00E-06(1.12E-06) & 1.08E-06(3.88E-06)$\approx$ & \cellcolor[rgb]{.851,.851,.851}1.00E-06(7.56E-07)$\approx$ & 1.13E-01(1.04E+00)+ & 5.06E-05(9.10E-05)+ & 1.24E-06(6.59E-07)$\approx$ & 2.97E-06(7.95E-06)+ & \cellcolor[rgb]{.851,.851,.851}1.00E-06(0.00E+00)$\approx$ \\
    \midrule
    \multirow{2}[2]{*}{SMD10} & $Acc_u$  & 2.73E+00(1.60E+01) & 1.60E+01(2.52E-07)$\approx$ & 1.60E+01(3.57E-08)$\approx$ & 3.86E-02(6.38E-02)$\approx$ & \cellcolor[rgb]{.851,.851,.851}4.84E-03(4.51E-01)$\approx$ & 1.60E+01(5.18E-04)+ & 3.97E+00(5.85E-02)+ & 3.17E+00(4.63E+00)+ \\
          & $Acc_l$  & 5.00E-06(3.00E+00) & 3.25E-06(1.60E+01)$\approx$ & \cellcolor[rgb]{.851,.851,.851}1.00E-06(4.15E-01)- & 3.24E-02(4.99E-02)$\approx$ & 4.84E-03(1.48E+00)$\approx$ & 1.60E+01(1.55E-03)+ & 8.71E-01(8.24E-01)+ & 6.98E-01(6.25E-01)+ \\
    \midrule
    \multirow{2}[2]{*}{SMD11} & $Acc_u$  & 8.62E-04(1.96E-03) & 6.99E-04(1.29E-03)$\approx$ & 1.13E-03(2.76E-03)+ & 4.45E-03(6.82E-03)+ & 1.52E+01(3.19E+01)+ & 4.32E-04(6.69E-04)- & 1.00E+00(1.16E-04)+ & \cellcolor[rgb]{.851,.851,.851}1.00E-06(9.98E-01)- \\
          & $Acc_l$  & 1.83E-03(1.90E-03) & 9.64E-04(2.41E-03)- & 2.54E-03(5.23E-03)$\approx$ & 9.52E-03(1.27E-02)+ & 1.64E+01(3.37E+01)+ & 4.58E-04(7.06E-04)- & 1.00E+00(1.16E-04)+ & \cellcolor[rgb]{.851,.851,.851}1.00E-06(7.25E-01)- \\
    \midrule
    \multirow{2}[2]{*}{SMD12} & $Acc_u$  & \cellcolor[rgb]{.851,.851,.851}1.00E-06(0.00E+00) & \cellcolor[rgb]{.851,.851,.851}1.00E-06(0.00E+00)$\approx$ & \cellcolor[rgb]{.851,.851,.851}1.00E-06(0.00E+00)$\approx$ & 3.64E-02(4.38E-02)+ & 8.19E-02(1.45E-01)+ & 5.11E-04(5.50E-04)+ & 6.86E+00(3.00E-06)+ & 3.66E+00(3.57E+00)+ \\
          & $Acc_l$  & \cellcolor[rgb]{.851,.851,.851}4.56E-06(1.60E+01) & 9.26E+00(1.60E+01)+ & 2.12E-05(2.13E-03)+ & 3.92E-02(5.49E-02)+ & 1.79E-01(5.34E-01)+ & 1.60E+01(1.65E-03)+ & 2.10E-01(1.98E-02)+ & 1.31E+00(1.20E+00)+ \\
\midrule
     \multicolumn{3}{l}
     {+/-/$\approx$ ($Acc_u$ \& $Acc_l$)} &1/1/10 \& 3/1/8 & 2/1/9 \& 4/1/7 &10/0/2 \& 10/0/2 & 6/0/6 \& 6/0/6 & 4/1/7 \& 5/2/5 & 12/0/0 \& 10/2/0 & 2/2/8 \& 3/3/6 \\
    \bottomrule
    \end{tabular}
    }
     \begin{tablenotes}
        \footnotesize
        \item "+", "$\approx$" and "-" indicate that DRC-CMA-ES performs significantly better or equivalently or worse than the compared algorithms at a 0.05 significance level by Wilcoxon test, respectively. The same symbols applies to other tables.
      \end{tablenotes}
\end{table*}

Notably, the two indicators should not be considered in isolation. When algorithms achieve similar accuracy, the number of FEs can reflect the differences in algorithm efficiency.
However, if an algorithm achieves significantly worse accuracy than others, even if it consumes less resources, it cannot be considered efficient.
This is because it may fall into premature convergence, revealing limitations in its fundamental problem-solving capability.

\vspace{-0.5em}
\subsection{Compared Algorithms and Parameter Settings}
To verify the effectiveness of DRC-CMA-ES, seven well-regarded evolutionary bilevel optimization algorithms are selected for comparison: 

$\bullet$ BL-CMA-ES \cite{he2018evolutionary}: a covariance matrix adaptation-based BLEA that employs CMA-ES at both levels, with prior knowledge about the distribution shared from the upper level to the lower level.

$\bullet$ TLEA-CMA-ES \cite{chen2021transfer}: a transfer learning-based BLEA that optimizes a set of lower-level tasks in parallel, and a knowledge transfer strategy is designed.

$\bullet$ NBLEA \cite{sinha2014test}: a nested BLEA that optimizes the upper and lower levels separately using two steady-state GA optimizers.

$\bullet$ BLEAQ-II \cite{sinha2020bilevel}: a quadratic approximation-based BLEA that employs the lower-level rational reaction mapping and the optimal function value mapping in a nested structure.

$\bullet$ GO-CMA-ES \cite{huang2020framework}: a two-stage BLEA that first groups variables based on the interactions between upper- and lower-level variables, then optimizes the subgroups.

$\bullet$ MFBLA \cite{mamun2021multifidelity}: a multifidelity BLEA that adaptively assigns different fidelities to evaluate solutions during the search, instead of resorting to exhaustive lower-level optimization.

$\bullet$ SA-BL-IMODE \cite{lin2023classification}: a surrogate-assisted BLEA that integrates preselection and environment selection strategies with a classification model.

The parameters for the test environments and algorithms are set as follows.
For algorithms using CMA-ES as the optimizer, the upper-level population size $p$ and the lower-level population size $q$ are set to $4+\lfloor\ln(m+n)\rfloor$ and $4+\lfloor\ln(n)\rfloor$, respectively.
Other parameters of CMA-ES remain unchanged from the default settings in \cite{hansen2001completely}.
Lower-level optimization terminates if 1) the number of consumed lower-level FEs exceeds $FEs_l^{max}$ or 2) the elitist lower-level objective values vary by less than $1e^{-5}$ over the last $FEs_l^{var}$. 
Upper-level optimization terminates if 1) the number of consumed upper-level FEs exceeds $FEs_u^{max}$ or 2) the elitist upper-level objective values vary by less than $1e^{-6}$ over the last $FEs_u^{var}$ or 3) the accuracy of the elitist upper-level objective value is less than $1e^{-6}$.
The detailed parameter settings for the termination criteria are listed in Table ~\ref{table:termination criteria}.
For other algorithms, the population size at both levels is set to 50, and other parameters are referenced from the original papers.

All experiments were implemented in MATLAB and executed on a computer with a 3.80 GHz Intel Xeon(R) Gold 5222 CPU.
Each algorithm was run 21 times independently on each test instance.

\vspace{-0.5em}
\subsection{Comparison Study} 
The statistical results of the median and interquartile range (IQR) of the accuracy and number of FEs achieved by all algorithms in the comparison experiment on SMD are presented in {Tables~\ref{table:Acc_m2n3}, ~\ref{table:FEul_m2n3} and ~\ref{table:FEs}.
For an intuitive comparison, the best result in each case is highlighted with a gray background, and the Wilcoxon test was conducted at a 0.05 significance level to indicate the difference between comparison results.

\begin{table*}[htbp]
  \centering
  \caption{Median and IQR results of number of $FEs_u$ and $FEs_l$ on SMD problems with dimension  ($m$ = 2, $n$ = 3)}
\label{table:FEul_m2n3}
\resizebox{\textwidth}{!}{
\begin{tabular}{cccccccccc}
    \toprule
    \multicolumn{2}{c}{($m$ = 2, $n$ = 3)} & DRC-CMA-ES (ours) & BL-CMA-ES & TLEA-CMA-ES & NBLEA & BLEAQ-II & GO-CMA-ES & MFBLA & SA-BL-IMODE \\
    \midrule
    \multirow{2}[0]{*}{SMD1} & $FEs_u$  & 8.18E+02(1.09E+02) & 3.15E+02(2.30E+01)- & 3.15E+02(3.30E+01)- & 1.22E+03(2.30E+02)+ & 7.76E+02(5.30E+02)$\approx$ & 5.99E+02(3.20E+01)- & 8.48E+02(7.90E+01)+ & \cellcolor[rgb]{.851,.851,.851}1.48E+02(3.80E+01)- \\
          & $FEs_l$  & \cellcolor[rgb]{.851,.851,.851}1.37E+04(1.52E+03) & 2.04E+04(1.79E+03)+ & 2.04E+04(1.22E+03)+ & 5.22E+05(8.47E+04)+ & 5.37E+04(3.35E+04)+ & 1.87E+04(1.82E+03)+ & 7.47E+04(6.20E+03)+ & 2.21E+04(4.96E+03)+ \\
    \midrule
    \multirow{2}[0]{*}{SMD2} & $FEs_u$  & 7.74E+02(5.70E+01) & 2.80E+02(3.70E+01)- & 2.97E+02(3.40E+01)- & 1.42E+03(3.72E+02)+ & 1.05E+03(6.32E+02)+ & 5.86E+02(2.00E+01)- & 8.86E+02(1.09E+02)+ & \cellcolor[rgb]{.851,.851,.851}1.21E+02(5.20E+01)- \\
          & $FEs_l$  & \cellcolor[rgb]{.851,.851,.851}1.28E+04(8.75E+02) & 1.96E+04(2.26E+03)+ & 2.00E+04(2.34E+03)+ & 6.18E+05(1.33E+05)+ & 7.26E+04(2.80E+04)+ & 1.80E+04(1.18E+03)+ & 8.00E+04(1.30E+04)+ & 1.80E+04(6.95E+03)+ \\
    \midrule
    \multirow{2}[0]{*}{SMD3} & $FEs_u$  & 8.35E+02(6.20E+01) & 3.27E+02(2.90E+01)- & 3.27E+02(2.10E+01)- & 1.24E+03(2.02E+02)+ & 6.11E+02(4.84E+02)- & 5.94E+02(3.10E+01)- & 8.73E+02(5.60E+01)+ & \cellcolor[rgb]{.851,.851,.851}1.33E+02(3.10E+01)- \\
          & $FEs_l$  & \cellcolor[rgb]{.851,.851,.851}1.43E+04(1.02E+03) & 2.13E+04(1.72E+03)+ & 2.07E+04(1.44E+03)+ & 5.81E+05(6.49E+04)+ & 6.16E+04(2.24E+04)+ & 2.11E+04(1.66E+03)+ & 7.70E+04(3.93E+03)+ & 2.30E+04(4.52E+03)+ \\
    \midrule
    \multirow{2}[0]{*}{SMD4} & $FEs_u$  & 8.82E+02(1.00E+02) & 3.39E+02(5.20E+01)- & 3.22E+02(5.10E+01)- & 1.40E+03(4.26E+02)+ & 9.07E+02(6.30E+02)$\approx$ & 5.67E+02(4.60E+01)- & 9.85E+02(8.90E+01)+ & \cellcolor[rgb]{.851,.851,.851}1.29E+02(3.50E+01)- \\
          & $FEs_l$  & \cellcolor[rgb]{.851,.851,.851}1.47E+04(1.95E+03) & 2.20E+04(3.09E+03)+ & 2.10E+04(1.54E+03)+ & 5.66E+05(1.26E+05)+ & 5.41E+04(3.05E+04)+ & 1.90E+04(1.98E+03)+ & 8.57E+04(1.07E+04)+ & 1.94E+04(4.70E+03)+ \\
    \midrule
    \multirow{2}[0]{*}{SMD5} & $FEs_u$  & 8.73E+02(1.35E+02) & 3.53E+02(6.10E+01)- & 3.39E+02(5.60E+01)- & 1.39E+03(6.31E+02)+ & 1.27E+03(1.07E+03)+ & 6.02E+02(5.50E+01)- & 9.69E+02(1.84E+02)+ & \cellcolor[rgb]{.851,.851,.851}1.99E+02(8.40E+01)- \\
          & $FEs_l$  & \cellcolor[rgb]{.851,.851,.851}1.48E+04(1.75E+03) & 2.16E+04(1.93E+03)+ & 2.04E+04(3.13E+03)+ & 6.15E+05(2.78E+05)+ & 8.55E+04(4.48E+04)+ & 1.82E+04(1.67E+03)+ & 8.65E+04(1.86E+04)+ & 3.10E+04(1.06E+04)+ \\
    \midrule
    \multirow{2}[0]{*}{SMD6} & $FEs_u$  & 1.20E+03(1.73E+02) & 4.33E+02(4.80E+01)- & 4.28E+02(5.00E+01)- & 1.77E+03(7.25E+02)+ & 3.90E+02(7.34E+02)- & 9.16E+02(4.80E+01)- & 1.00E+03(7.40E+01)- & \cellcolor[rgb]{.851,.851,.851}1.62E+02(6.40E+01)- \\
          & $FEs_l$  & 1.97E+04(2.63E+03) & 2.40E+04(2.51E+03)+ & 2.33E+04(2.85E+03)+ & 3.32E+04(1.34E+04)+ & \cellcolor[rgb]{.851,.851,.851}3.06E+03(4.58E+03)- & 2.06E+04(1.33E+03)+ & 8.72E+04(5.00E+03)+ & 2.17E+04(8.23E+03)+ \\
    \midrule
    \multirow{2}[0]{*}{SMD7} & $FEs_u$  & 8.48E+02(7.83E+02) & 5.29E+02(3.63E+02)- & 6.37E+02(4.18E+02)- & 1.33E+03(3.23E+02)+ & 1.36E+03(4.82E+02)+ & 8.13E+02(6.60E+01)- & 8.97E+02(7.90E+01)+ & \cellcolor[rgb]{.851,.851,.851}1.33E+02(9.80E+01)- \\
          & $FEs_l$  & \cellcolor[rgb]{.851,.851,.851}1.39E+04(9.42E+03) & 2.34E+04(6.23E+03)+ & 2.59E+04(9.46E+03)+ & 5.46E+05(1.14E+05)+ & 8.78E+04(2.28E+04)+ & 1.77E+04(1.82E+03)+ & 7.95E+04(9.14E+03)+ & 1.93E+04(1.17E+04)+ \\
    \midrule
    \multirow{2}[0]{*}{SMD8} & $FEs_u$  & 2.51E+03(1.20E+01) & 1.29E+03(2.91E+02)- & 1.31E+03(1.87E+02)- & 1.83E+03(7.24E+02)- & 2.06E+03(9.58E+02)$\approx$ & 1.44E+03(1.22E+02)- & 9.27E+02(3.70E+01)- & \cellcolor[rgb]{.851,.851,.851}1.40E+02(8.70E+01)- \\
          & $FEs_l$  & 3.96E+04(1.45E+03) & 6.22E+04(1.12E+04)+ & 6.41E+04(1.01E+04)+ & 8.09E+05(3.01E+05)+ & 1.24E+05(5.26E+04)+ & \cellcolor[rgb]{.851,.851,.851}1.92E+04(1.25E+03)- & 8.35E+04(1.11E+03)+ & 2.29E+04(1.51E+04)- \\
    \midrule
    \multirow{2}[0]{*}{SMD9} & $FEs_u$  & 8.04E+02(1.34E+02) & 3.10E+02(7.10E+01)- & 3.28E+02(3.50E+01)- & 3.86E+03(4.45E+03)+ & 1.69E+03(5.77E+02)+ & 3.28E+02(7.10E+01)- & 9.54E+02(1.27E+02)+ & \cellcolor[rgb]{.851,.851,.851}1.22E+02(3.90E+01)- \\
          & $FEs_l$  & \cellcolor[rgb]{.851,.851,.851}1.35E+04(1.76E+03) & 1.98E+04(3.40E+03)+ & 2.00E+04(1.60E+03)+ & 1.92E+06(2.10E+06)+ & 1.22E+05(1.71E+04)+ & 2.05E+04(3.05E+03)+ & 8.42E+04(1.16E+04)+ & 2.08E+04(6.15E+03)+ \\
    \midrule
    \multirow{2}[0]{*}{SMD10} & $FEs_u$  & 2.51E+03(8.00E+00) & 1.55E+03(4.47E+02)- & 1.65E+03(1.66E+02)- & 2.29E+03(8.99E+02)$\approx$ & 1.66E+03(4.72E+02)- & 1.26E+03(7.10E+01)- & 8.67E+02(1.56E+02)- & \cellcolor[rgb]{.851,.851,.851}7.64E+02(1.15E+03)- \\
          & $FEs_l$  & \cellcolor[rgb]{.851,.851,.851}4.03E+04(6.58E+02) & 6.83E+04(1.41E+04)+ & 7.30E+04(7.71E+03)+ & 1.58E+06(5.51E+05)+ & 1.04E+05(3.17E+04)+ & 4.70E+04(2.51E+03)+ & 7.60E+04(1.36E+04)+ & 1.60E+05(2.45E+05)+ \\
    \midrule
    \multirow{2}[0]{*}{SMD11} & $FEs_u$  & 2.51E+03(1.20E+01) & 2.50E+03(1.98E+02)$\approx$ & 2.50E+03(8.47E+02)$\approx$ & 2.12E+03(1.21E+03)- & 4.37E+03(4.81E+03)+ & 2.50E+03(2.91E+02)$\approx$ & 8.70E+02(6.10E+01)- & \cellcolor[rgb]{.851,.851,.851}4.23E+02(8.09E+02)- \\
          & $FEs_l$  & \cellcolor[rgb]{.851,.851,.851}4.27E+04(7.42E+02) & 1.27E+05(2.24E+04)+ & 1.22E+05(4.40E+04)+ & 1.31E+06(7.00E+05)+ & 1.77E+06(2.24E+06)+ & 8.00E+04(7.65E+03)+ & 7.67E+04(6.51E+03)+ & 6.25E+04(1.22E+05)+ \\
    \midrule
    \multirow{2}[1]{*}{SMD12} & $FEs_u$  & 2.14E+03(2.74E+02) & 9.32E+02(1.03E+02)- & 8.26E+02(2.03E+02)- & 2.15E+03(1.02E+03)$\approx$ & 1.70E+03(9.61E+02)- & 1.19E+03(1.00E+02)- & 8.62E+02(6.10E+01)- & \cellcolor[rgb]{.851,.851,.851}5.94E+02(4.31E+02)- \\
          & $FEs_l$  & \cellcolor[rgb]{.851,.851,.851}3.40E+04(3.62E+03) & 4.69E+04(3.72E+03)+ & 4.46E+04(7.74E+03)+ & 1.68E+06(7.22E+05)+ & 2.22E+05(9.89E+04)+ & 4.78E+04(4.81E+03)+ & 7.59E+04(6.33E+03)+ & 1.37E+05(9.43E+04)+ \\
    \midrule
     \multicolumn{3}{l}
     {+/-/$\approx$ ($FEs_u$ \& $FEs_l$)} &0/11/1 \& 12/0/0 &0/11/1 \& 12/0/0 &8/2/2 \& 12/0/0 & 5/4/3 \& 11/1/0 &  0/11/1 \& 11/1/0 & 7/5/0 \& 12/0/0 & 0/12/0 \& 11/1/0\\
      \bottomrule
    \end{tabular}%
    }
\end{table*}%

As shown in Table~\ref{table:Acc_m2n3}, DRC-CMA-ES achieves the best upper-level accuracy in 9 out of 12 SMD problems and the best lower-level accuracy in 8 out of 12 SMD problems with dimension ($m$ = 2, $n$ = 3).
In the total 24 results of upper- and lower-level accuracy,
DRC-CMA-ES and SA-BL-IMODE obtain 17 and 18 best results, respectively, making them the top two algorithms.
Statistically, the Wilcoxon test results indicate that there is no significant difference between their performances.
Compared to other algorithms, DRC-CMA-ES performs slightly better than the other three CMA-ES-based algorithms and significantly outperforms the remaining algorithms.
Among these, BLEAQ-II outperforms NBLEA due to the approximation between upper and lower levels, but the improvement is limited because most problems in SMD test suite are not linear or quadratic.

Note that, although DRC-CMA-ES suggests not to perform complete optimization for each lower-level task, the most promising tasks in the quasi-parallel evolution process are prioritized through the dynamic resource competition mechanism.
As a result, the accuracy of DRC-CMA-ES can be guaranteed, which achieves or approaches $1e^{-6}$ on most SMD instances. 
In addition, the small interquartile range observed in repeated experiments indicates that DRC-CMA-ES converges stably.

The experiment on SMD with dimension  ($m$ = 10, $n$ = 10) and ($m$ = 30, $n$ = 30) were conducted to investigate the scalability of the algorithms, and the results are summarized in Tables S-I and S-II in the supplementary material.
As the search space expanded, the accuracy of all algorithms decreased significantly compared to the lower-dimensional cases. This decrease was especially pronounced for approximation-based algorithms, as their model accuracy is challenged by the increased number of variables.
However, the advantage of DRC-CMA-ES became more apparent in high-dimensional settings, as it still obtains the best results on most instances.
The comparison results indicate that DRC-CMA-ES can achieve competitive accuracy across various problem scales.

\begin{table*}[htbp]
  \centering
  \caption{Median results of the number of total $FEs$ on SMD problems on different scales}
\label{table:FEs}
\resizebox{\textwidth}{!}{
\begin{tabular}{cccccccccc}
\toprule
    Prob  & Dimension & DRC-CMA-ES (ours) & BL-CMA-ES & TLEA-CMA-ES & NBLEA & BLEAQ-II & GO-CMA-ES & MFBLA & SA-BL-IMODE \\
    \midrule
    \multirow{3}[2]{*}{SMD1} & ($m$ = 2, $n$ =3) & \cellcolor[rgb]{.851,.851,.851}1.46E+04 & 2.07E+04(29.8\%) & 2.07E+04(29.7\%) & 5.23E+05(97.2\%) & 5.44E+04(73.2\%) & 1.93E+04(24.5\%) & 7.55E+04(80.7\%) & 2.22E+04(34.4\%) \\
          & ($m$ = 10, $n$ =10) & \cellcolor[rgb]{.851,.851,.851}9.95E+04 & 2.05E+05(51.6\%) & 1.88E+05(47.2\%) & 7.96E+05(87.5\%) & 6.57E+05(84.8\%) & 2.89E+05(65.5\%) & 6.91E+05(85.6\%) & 4.43E+05(77.6\%) \\
          & ($m$ = 30, $n$ =30) & \cellcolor[rgb]{.851,.851,.851}3.29E+05 & 6.42E+05(48.7\%) & 6.49E+05(49.3\%) & 3.16E+06(89.6\%) & 5.61E+06(94.1\%) & 6.84E+05(51.9\%) & 2.19E+06(85.0\%) & 1.16E+06(71.5\%) \\
    \midrule
    \multirow{3}[2]{*}{SMD2} & ($m$ = 2, $n$ =3) & \cellcolor[rgb]{.851,.851,.851}1.36E+04 & 1.99E+04(31.7\%) & 2.03E+04(33.0\%) & 6.20E+05(97.8\%) & 7.37E+04(81.6\%) & 1.86E+04(27.0\%) & 8.08E+04(83.2\%) & 1.81E+04(24.8\%) \\
          & ($m$ = 10, $n$ =10) & \cellcolor[rgb]{.851,.851,.851}9.86E+04 & 2.01E+05(50.9\%) & 1.84E+05(46.4\%) & 7.39E+05(86.7\%) & 6.86E+05(85.6\%) & 2.82E+05(65.1\%) & 7.14E+05(86.2\%) & 2.37E+05(58.4\%) \\
          & ($m$ = 30, $n$ =30) & \cellcolor[rgb]{.851,.851,.851}3.19E+05 & 6.54E+05(51.2\%) & 6.52E+05(51.0\%) & 1.90E+06(83.2\%) & 3.38E+06(90.6\%) & 6.84E+05(53.4\%) & 2.15E+06(85.1\%) & 6.45E+05(50.5\%) \\
    \midrule
    \multirow{3}[2]{*}{SMD3} & ($m$ = 2, $n$ =3) & \cellcolor[rgb]{.851,.851,.851}1.52E+04 & 2.16E+04(29.7\%) & 2.10E+04(27.9\%) & 5.82E+05(97.4\%) & 6.27E+04(75.8\%) & 2.16E+04(29.9\%) & 7.79E+04(80.5\%) & 2.31E+04(34.4\%) \\
          & ($m$ = 10, $n$ =10) & \cellcolor[rgb]{.851,.851,.851}1.02E+05 & 1.68E+05(39.7\%) & 1.73E+05(41.2\%) & 8.76E+05(88.4\%) & 8.59E+05(88.2\%) & 2.94E+05(65.5\%) & 7.76E+05(86.9\%) & 4.52E+05(77.5\%) \\
          & ($m$ = 30, $n$ =30) & \cellcolor[rgb]{.851,.851,.851}3.40E+05 & 5.22E+05(34.9\%) & 5.40E+05(37.0\%) & 3.62E+06(90.6\%) & 5.81E+06(94.1\%) & 7.05E+05(51.8\%) & 2.23E+06(84.7\%) & 1.17E+06(71.0\%) \\
    \midrule
    \multirow{3}[2]{*}{SMD4} & ($m$ = 2, $n$ =3) & \cellcolor[rgb]{.851,.851,.851}1.56E+04 & 2.24E+04(30.1\%) & 2.13E+04(26.8\%) & 5.67E+05(97.2\%) & 5.53E+04(71.7\%) & 1.96E+04(20.1\%) & 8.67E+04(82.0\%) & 1.96E+04(20.2\%) \\
          & ($m$ = 10, $n$ =10) & \cellcolor[rgb]{.851,.851,.851}1.15E+05 & 1.96E+05(41.1\%) & 1.95E+05(40.7\%) & 8.18E+05(85.9\%) & 2.45E+05(52.9\%) & 2.84E+05(59.3\%) & 7.24E+05(84.1\%) & 1.35E+05(14.5\%) \\
          & ($m$ = 30, $n$ =30) & \cellcolor[rgb]{.851,.851,.851}4.22E+05 & 7.18E+05(41.2\%) & 6.81E+05(38.0\%) & 1.13E+06(62.6\%) & 2.69E+06(84.3\%) & 7.00E+05(39.6\%) & 2.20E+06(80.8\%) & 4.58E+05(7.9\%) \\
    \midrule
    \multirow{3}[2]{*}{SMD5} & ($m$ = 2, $n$ =3) & \cellcolor[rgb]{.851,.851,.851}1.56E+04 & 2.19E+04(28.7\%) & 2.07E+04(24.5\%) & 6.17E+05(97.5\%) & 8.66E+04(81.9\%) & 1.88E+04(17.0\%) & 8.75E+04(82.1\%) & 3.12E+04(49.8\%) \\
          & ($m$ = 10, $n$ =10) & \cellcolor[rgb]{.851,.851,.851}1.37E+05 & 2.37E+05(42.0\%) & 2.25E+05(39.0\%) & 8.19E+05(83.2\%) & 6.07E+05(77.4\%) & 2.87E+05(52.1\%) & 8.51E+05(83.9\%) & 6.43E+05(78.6\%) \\
          & ($m$ = 30, $n$ =30) & \cellcolor[rgb]{.851,.851,.851}4.38E+05 & 7.89E+05(44.5\%) & 7.74E+05(43.5\%) & 1.68E+06(74.0\%) & 2.98E+06(85.3\%) & 6.76E+05(35.3\%) & 2.39E+06(81.7\%) & 1.29E+06(66.2\%) \\
    \midrule
    \multirow{3}[2]{*}{SMD6} & ($m$ = 2, $n$ =3) & 2.09E+04 & 2.44E+04(14.4\%) & 2.38E+04(12.1\%) & 3.49E+04(40.2\%) & \cellcolor[rgb]{.851,.851,.851}3.45E+03(-83.5\%) & 2.15E+04(2.9\%) & 8.82E+04(76.3\%) & 2.19E+04(4.5\%) \\
          & ($m$ = 10, $n$ =10) & 1.20E+05 & 2.30E+05(47.9\%) & 2.09E+05(42.8\%) & 5.06E+05(76.4\%) & \cellcolor[rgb]{.851,.851,.851}5.49E+04(-54.0\%) & 2.41E+05(50.5\%) & 7.57E+05(84.2\%) & 2.78E+05(57.0\%) \\
          & ($m$ = 30, $n$ =30) & 3.90E+05 & 7.45E+05(47.6\%) & 7.49E+05(47.9\%) & 2.76E+06(85.8\%) & 2.45E+06(84.1\%) & 5.99E+05(34.8\%) & 2.23E+06(82.5\%) & \cellcolor[rgb]{.851,.851,.851}2.65E+05(-32.2\%) \\
    \midrule
    \multirow{3}[1]{*}{SMD7} & ($m$ = 2, $n$ =3) & \cellcolor[rgb]{.851,.851,.851}1.48E+04 & 2.39E+04(38.3\%) & 2.65E+04(44.3\%) & 5.48E+05(97.3\%) & 8.97E+04(83.6\%) & 1.85E+04(20.3\%) & 8.04E+04(81.6\%) & 1.94E+04(24.0\%) \\
          & ($m$ = 10, $n$ =10) & \cellcolor[rgb]{.851,.851,.851}1.26E+05 & 1.87E+05(32.7\%) & 1.93E+05(34.7\%) & 9.00E+05(86.0\%) & 7.59E+05(83.4\%) & 2.64E+05(52.2\%) & 8.31E+05(84.8\%) & 4.09E+05(69.1\%) \\
          & ($m$ = 30, $n$ =30) & \cellcolor[rgb]{.851,.851,.851}3.53E+05 & 1.47E+06(76.0\%) & 1.47E+06(76.0\%) & 1.60E+06(77.9\%) & 3.08E+06(88.5\%) & 3.84E+05(8.1\%) & 2.32E+06(84.8\%) & 8.15E+05(56.6\%) \\
    \midrule
    \multirow{3}[0]{*}{SMD8} & ($m$ = 2, $n$ =3) & 4.21E+04 & 6.35E+04(33.6\%) & 6.54E+04(35.5\%) & 8.11E+05(94.8\%) & 1.26E+05(66.6\%) & \cellcolor[rgb]{.851,.851,.851}2.06E+04(-51.0\%) & 8.44E+04(50.1\%) & 2.30E+04(-45.4\%) \\
          & ($m$ = 10, $n$ =10) & \cellcolor[rgb]{.851,.851,.851}1.55E+05 & 6.81E+05(77.3\%) & 6.56E+05(76.4\%) & 8.36E+05(81.5\%) & 9.55E+05(83.8\%) & 2.62E+05(41.0\%) & 9.17E+05(83.1\%) & 6.78E+05(77.2\%) \\
          & ($m$ = 30, $n$ =30) & \cellcolor[rgb]{.851,.851,.851}4.35E+05 & 1.41E+06(69.1\%) & 1.46E+06(70.2\%) & 1.96E+06(77.8\%) & 3.25E+06(86.6\%) & 6.78E+05(35.8\%) & 2.22E+06(80.4\%) & 1.50E+06(71.1\%) \\
    \midrule
    \multirow{3}[1]{*}{SMD9} & ($m$ = 2, $n$ =3) & \cellcolor[rgb]{.851,.851,.851}1.43E+04 & 2.01E+04(28.7\%) & 2.04E+04(29.6\%) & 1.92E+06(99.3\%) & 1.24E+05(88.4\%) & 2.08E+04(31.1\%) & 8.52E+04(83.2\%) & 2.09E+04(31.5\%) \\
          & ($m$ = 10, $n$ =10) & \cellcolor[rgb]{.851,.851,.851}1.43E+05 & 2.30E+05(38.0\%) & 2.10E+05(31.9\%) & 9.43E+05(84.9\%) & 3.75E+05(61.9\%) & 2.27E+05(37.2\%) & 7.06E+05(79.8\%) & 2.29E+05(37.6\%) \\
          & ($m$ = 30, $n$ =30) & \cellcolor[rgb]{.851,.851,.851}4.31E+05 & 1.43E+06(69.9\%) & 1.64E+06(73.7\%) & 1.95E+06(77.9\%) & 3.45E+06(87.5\%) & 1.70E+06(74.6\%) & 2.32E+06(81.4\%) & 5.62E+05(23.3\%) \\
    \midrule
    \multirow{3}[2]{*}{SMD10} & ($m$ = 2, $n$ =3) & \cellcolor[rgb]{.851,.851,.851}4.28E+04 & 6.98E+04(38.7\%) & 7.46E+04(42.7\%) & 1.58E+06(97.3\%) & 1.06E+05(59.5\%) & 4.83E+04(11.4\%) & 7.68E+04(44.3\%) & 1.61E+05(73.4\%) \\
          & ($m$ = 10, $n$ =10) & \cellcolor[rgb]{.851,.851,.851}1.59E+05 & 3.96E+05(59.8\%) & 3.97E+05(59.9\%) & 5.00E+06(96.8\%) & 4.55E+05(65.0\%) & 3.97E+05(59.9\%) & 7.71E+05(79.3\%) & 7.15E+05(77.7\%) \\
          & ($m$ = 30, $n$ =30) & \cellcolor[rgb]{.851,.851,.851}4.64E+05 & 1.09E+06(57.5\%) & 1.10E+06(57.8\%) & 8.08E+06(94.3\%) & 3.26E+06(85.8\%) & 1.05E+06(55.7\%) & 2.31E+06(80.0\%) & 2.70E+06(82.8\%) \\
    \midrule
    \multirow{3}[2]{*}{SMD11} & ($m$ = 2, $n$ =3) & \cellcolor[rgb]{.851,.851,.851}4.52E+04 & 1.29E+05(65.1\%) & 1.25E+05(63.7\%) & 1.31E+06(96.5\%) & 1.78E+06(97.5\%) & 8.25E+04(45.2\%) & 7.76E+04(41.8\%) & 6.29E+04(28.2\%) \\
          & ($m$ = 10, $n$ =10) & \cellcolor[rgb]{.851,.851,.851}1.42E+05 & 3.50E+05(59.3\%) & 4.12E+05(65.4\%) & 3.07E+06(95.4\%) & 5.88E+05(75.7\%) & 2.58E+05(44.9\%) & 7.07E+05(79.9\%) & 2.94E+05(51.5\%) \\
          & ($m$ = 30, $n$ =30) & \cellcolor[rgb]{.851,.851,.851}3.97E+05 & 1.04E+06(61.8\%) & 1.05E+06(62.2\%) & 5.51E+06(92.8\%) & 3.55E+06(88.8\%) & 6.73E+05(41.1\%) & 2.17E+06(81.7\%) & 5.46E+05(27.4\%) \\
    \midrule
    \multirow{3}[2]{*}{SMD12} & ($m$ = 2, $n$ =3) & \cellcolor[rgb]{.851,.851,.851}3.62E+04 & 4.79E+04(24.4\%) & 4.54E+04(20.4\%) & 1.69E+06(97.9\%) & 2.25E+05(83.9\%) & 4.90E+04(26.2\%) & 7.68E+04(52.9\%) & 1.37E+05(73.6\%) \\
          & ($m$ = 10, $n$ =10) & \cellcolor[rgb]{.851,.851,.851}1.56E+05 & 4.18E+05(62.6\%) & 4.12E+05(62.1\%) & 3.53E+06(95.6\%) & 4.67E+05(66.5\%) & 3.67E+05(57.4\%) & 7.54E+05(79.3\%) & 6.08E+05(74.3\%) \\
          & ($m$ = 30, $n$ =30) & \cellcolor[rgb]{.851,.851,.851}4.56E+05 & 1.18E+06(61.3\%) & 1.19E+06(61.7\%) & 1.05E+07(95.7\%) & 3.99E+06(88.6\%) & 9.84E+05(53.7\%) & 2.17E+06(79.0\%) & 1.75E+06(74.0\%) \\
    \midrule
    \multicolumn{1}{c}{\multirow{3}[2]{*}{SMD\newline{}-Avg}} & ($m$ = 2, $n$ =3) & \cellcolor[rgb]{.851,.851,.851}2.42E+04 & 4.05E+04(40.1\%) & 4.04E+04(40.0\%) & 9.00E+05(97.3\%) & 2.32E+05(89.6\%) & 2.99E+04(19.0\%) & 8.15E+04(70.3\%) & 4.67E+04(48.1\%) \\
          & ($m$ = 10, $n$ =10) & \cellcolor[rgb]{.851,.851,.851}1.29E+05 & 2.92E+05(55.6\%) & 2.88E+05(55.0\%) & 1.57E+06(91.7\%) & 5.59E+05(76.8\%) & 2.88E+05(55.0\%) & 7.67E+05(83.1\%) & 4.27E+05(69.6\%) \\
          & ($m$ = 30, $n$ =30) & \cellcolor[rgb]{.851,.851,.851}3.98E+05 & 9.74E+05(59.2\%) & 9.96E+05(60.1\%) & 3.65E+06(89.1\%) & 3.63E+06(89.0\%) & 7.93E+05(49.8\%) & 2.24E+06(82.2\%) & 1.07E+06(62.9\%) \\
   \bottomrule
    \end{tabular}
    }
    \begin{tablenotes}
        \footnotesize
        \item The values in parentheses indicate the relative percentage reduction of FEs that DRC-CMA-ES consumed, and SMD-Avg denotes the average value over all problems.
      \end{tablenotes}
\end{table*}%

The number of FEs of the comparison algorithms on SMD with dimension ($m$ = 2, $n$ = 3) is listed in Table~\ref{table:FEul_m2n3}.
It can be clearly observed that DRC-CMA-ES consumes the least computing resources for lower-level optimization in 10 out of 12 problems except SMD6 and SMD8, and the Wilcoxon test results indicate that the advantage of DRC-CMA-ES is significant.
SMD6 is characterized by an infinite number of optimal solutions at the lower level for any given upper-level individual.
With quadratic approximation, BLEAQ-II can quickly select one of the lower-level optima, thereby consuming less FEs.
In this case, the resource allocation mechanism of DRC-CMA-ES is disturbed by the infinite number of lower optimal solutions, but it still performs significantly better than the other competing algorithms.
As for SMD8, only GO-CMA-ES consumed fewer lower-level FEs than DRC-CMA-ES.
However, as can be observed from Table~\ref{table:Acc_m2n3}, GO-CMA-ES achieved relatively poor accuracy on this problem, indicating that its advantage in the number of lower-level FEs may be attribute to premature convergence.

The advantage of DRC-CMA-ES in lower-level computing resource consumption is attributed to the effective allocation of resources to the more promising ones among the parallel lower-level tasks, thereby reducing waste on lower-level tasks with low evolving potential.
In other words, the effective utilization rate of calling function evaluations is improved, which accelerates the convergence of the optimization process.

In terms of the number of upper-level FEs, SA-BL-IMODE outperforms the other algorithms.
The upper-level FEs consumed by DRC-CMA-ES is slightly higher than those of other CMA-ES-based algorithms.
This is not because DRC-CMA-ES requires more upper-level iterations, but rather because some of the upper-level function evaluations are invoked within the lower-level resource-competing mechanism to evaluate the tasks.

Considering the different strategies for consuming computing resources in comparison algorithms, it is more reasonable to observe the total number of FEs.
The median results of $FEs$ presented by all algorithms in 21 runs on different scales are summarized in Table \ref{table:FEs}.
It is evident that DRC-CMA-ES demonstrates overwhelmingly superior performance, as it consumes the fewest FEs on 32 out of 36 instances.
In contrast, NBLEA consumes the most FEs due to its complete nested structure. 
On most problems, the quadratic approximation model of BLEAQ-II is inconsistent with the problem characteristics, so the number of evaluations cannot be significantly reduced via effective approximation.
The reason for the poor results of MFBLA is the failure to strike a balance between accuracy and control of lower-level computing resources, while that of SA-BL-IMODE may be attributed to the uncertainty of the approximation model.

\begin{figure*}[htbp]
\setlength{\abovecaptionskip}{-0.3cm}
\includegraphics[width=\textwidth]{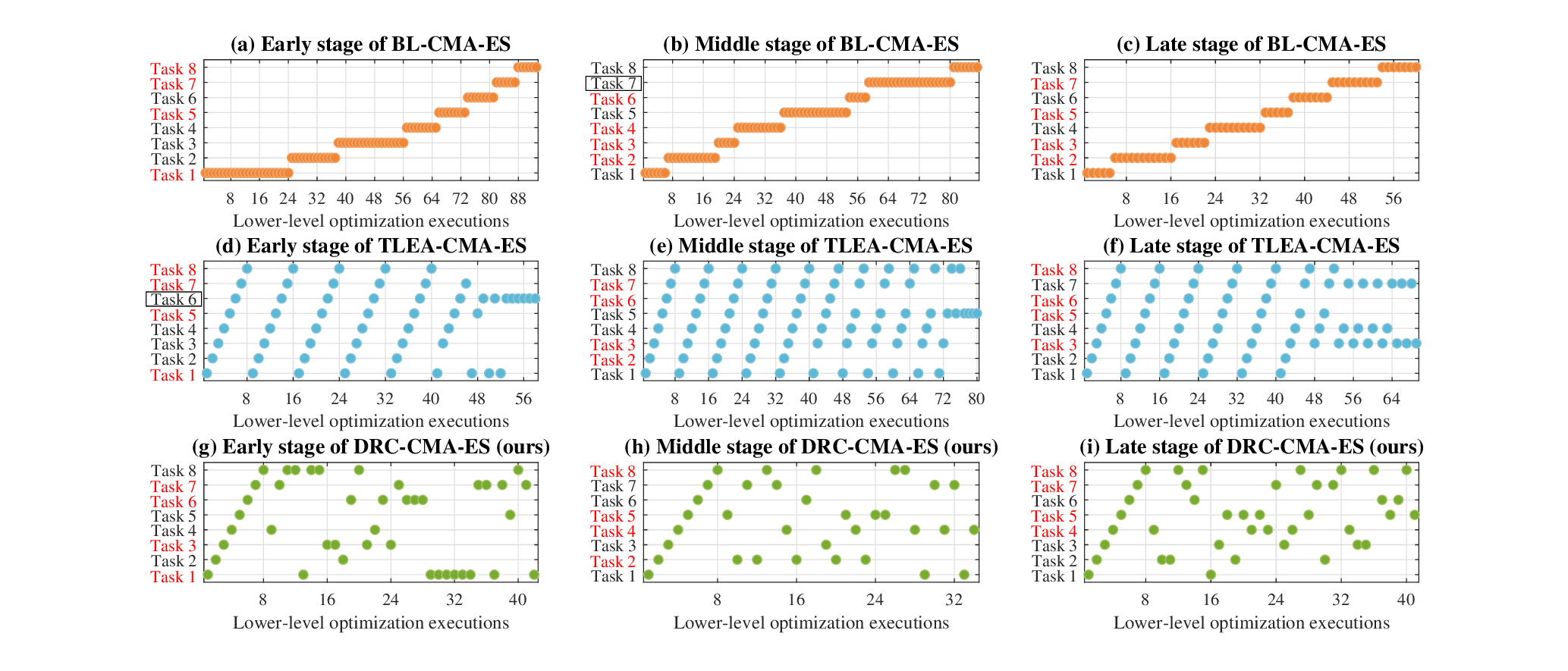}
\caption{The resource allocation of DRC-CMA-ES and the compared algorithms in the experiments on SMD1 with the median FEs results in 21 runs. The dot indicates that the lower-level task is selected and optimized for one lower-level iteration. The early, middle and late stages refer to the first upper-level iteration, and the 40\% and 80\% mark of the total upper-level iterations, respectively.  The tasks are derived from the currently generated upper-level individuals, so tasks with the same index in different sub-figures do not correspond to the same one. The tasks labeled in red indicate that the ($x_u$, $x_l^*$) pairs generated by these tasks survived the subsequent environmental selection. The tasks highlighted with a box are typical examples in which a large amount of lower-level FEs were executed, but the resulting pairs were not competitive enough and were eliminated.}
\label{fig:Resource_Allocation_compare}
\end{figure*}

\begin{figure*}[htbp]
\centering
\setlength{\abovecaptionskip}{-0.5cm}
\includegraphics[width=0.8\textwidth]{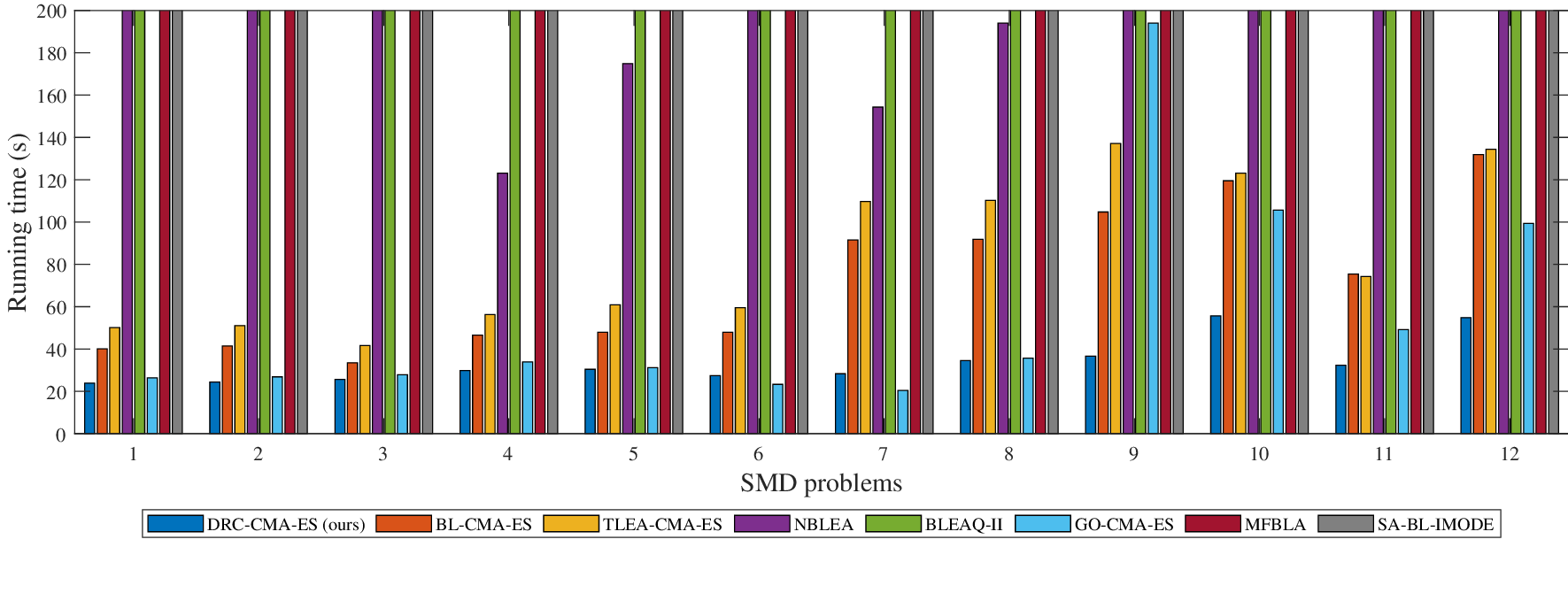}
\caption{Average running time of different algorithms on SMD problems with dimension ($m$ = 30, $n$ = 30). For ease of observation, only results within 200 seconds are presented.}
\label{fig:Running_time}
\end{figure*}

Although the performance of DRC-CMA-ES is slightly less efficient in terms of $FEs_u$ as seen in Table~\ref{table:FEul_m2n3}, it is superior in the total number of evaluations.
Actually, in a typical nested evolutionary framework, each newly generated lower-level vector for the upper-level individual needs to be evaluated. 
As a consequence, the total number of FEs consumed at the lower level is usually at least one order of magnitude higher than that at the upper level.
DRC-CMA-ES effectively reduces a large number of lower-level evaluations, providing a significant advantage in overall resource consumption performance.
As listed in parentheses in Table \ref{table:FEs}, DRC-CMA-ES consumes at least 30\% fewer computing resources than the comparison algorithms on most SMD problems, and up to 90\% fewer resources on some instances.
Benefiting from the resource allocation, the advantage of DRC-CMA-ES in computing efficiency is significantly pronounced as the dimension increases.

The comparison results and analysis on the TP test suite are provided in Section D of the supplementary material, indicaing that DRC-CMA-ES is applicable to various types of problems.
The experiments on the two test suites demonstrate that DRC-CMA-ES can significantly reduce the waste of computing resources, thereby improving the computational efficiency of bilevel optimization problems while achieving competitive accuracy.

\begin{table*}[htbp]
  \centering
 \caption{Median results of function values and number of FEs on Gold Mining and Decision Making problems}
    \label{table:GM and DM}  
     \resizebox{0.7\textwidth}{!}{
    \begin{tabular}{ccccccccc}
    \toprule
    GM    & DRC-CMA-ES (ours)& BL-CMA-ES & TLEA-CMA-ES & NBLEA & BLEAQ-II & GO-CMA-ES & MFBLA & SA-BL-IMODE \\
    \midrule
    $F$     & \cellcolor[rgb]{.851,.851,.851}-3.14E+02 & \cellcolor[rgb]{.851,.851,.851}-3.14E+02$\approx$ & \cellcolor[rgb]{.851,.851,.851}-3.14E+02$\approx$ & \cellcolor[rgb]{.851,.851,.851}-3.14E+02$\approx$ & \cellcolor[rgb]{.851,.851,.851}-3.14E+02$\approx$ & \cellcolor[rgb]{.851,.851,.851}-3.14E+02$\approx$ & \cellcolor[rgb]{.851,.851,.851}-3.14E+02$\approx$ & -3.11E+02+ \\
    $f$     & \cellcolor[rgb]{.851,.851,.851}1.32E+03 & \cellcolor[rgb]{.851,.851,.851}1.32E+03$\approx$ & 1.33E+03$\approx$ & \cellcolor[rgb]{.851,.851,.851}1.32E+03$\approx$ & \cellcolor[rgb]{.851,.851,.851}1.32E+03$\approx$ & \cellcolor[rgb]{.851,.851,.851}1.32E+03$\approx$ & 1.33E+03$\approx$ & 1.79E+03+ \\
    $FEs_u$  & 5.14E+02 & 5.43E+02$\approx$ & 3.64E+02- & 5.87E+02+ & 6.52E+02+ & 3.55E+02- & 9.89E+02+ & \cellcolor[rgb]{.851,.851,.851}1.39E+02- \\
    $FEs_l$  & \cellcolor[rgb]{.851,.851,.851}4.43E+03 & 1.26E+04+ & 8.51E+03+ & 1.94E+05+ & 5.34E+04+ & 8.57E+03+ & 2.98E+04+ & 1.70E+04+ \\
    $FEs$   & \cellcolor[rgb]{.851,.851,.851}4.94E+03 & 1.31E+04+ & 8.87E+03+ & 1.94E+05+ & 5.41E+04+ & 8.92E+03+ & 3.08E+04+ & 1.71E+04+ \\
    \midrule
    DM    & DRC-CMA-ES  (ours)& BL-CMA-ES & TLEA-CMA-ES & NBLEA & BLEAQ-II & GO-CMA-ES & MFBLA & SA-BL-IMODE \\
    \midrule
    $F$     & -1.08E+03 & \cellcolor[rgb]{.851,.851,.851}-1.09E+03$\approx$ & \cellcolor[rgb]{.851,.851,.851}-1.09E+03$\approx$ & -9.35E+02+ & \cellcolor[rgb]{.851,.851,.851}-1.09E+03$\approx$ & \cellcolor[rgb]{.851,.851,.851}-1.09E+03$\approx$ & -1.05E+03+ & -1.06E+03+ \\
    $f$     & \cellcolor[rgb]{.851,.851,.851}7.94E+03 & 8.79E+03$\approx$ & 1.06E+04+ & 8.33E+03$\approx$ & 1.07E+04+ & 8.07E+03+ & 1.10E+04+ & 8.17E+03+ \\
    $FEs_u$  & 2.50E+03 & 2.37E+03- & 2.50E+03$\approx$ & \cellcolor[rgb]{.851,.851,.851}4.56E+02- & 1.31E+03- & 1.60E+03- & 1.35E+03- & 8.67E+02- \\
    $FEs_l$  & \cellcolor[rgb]{.851,.851,.851}3.89E+04 & 1.09E+05+ & 1.15E+05+ & 1.38E+06+ & 5.59E+05+ & 6.99E+04+ & 1.13E+05+ & 2.53E+05+ \\
    $FEs$   & \cellcolor[rgb]{.851,.851,.851}4.14E+04 & 1.11E+05+ & 1.18E+05+ & 1.38E+06+ & 5.60E+05+ & 7.15E+04+ & 1.14E+05+ & 2.54E+05+ \\
    \bottomrule
\end{tabular}%
}
\end{table*}%

\vspace{-0.5em}
\subsection{Discussion on Resource Allocation} 
The dynamic resource allocation in the competitive quasi-parallel framework is the key characteristic that distinguishes DRC-CMA-ES from other bilevel evolutionary algorithms.
To further explore this mechanism, we present the resource allocation behavior of BL-CMA-ES, TLEA-CMA-ES and DRC-CMA-ES at different stages on SMD1 with dimension ($m$ = 2, $n$ = 3) in Fig. \ref{fig:Resource_Allocation_compare}.
As illustrated in Fig. \ref{fig:Resource_Allocation_compare}(a)-(f), BL-CMA-ES and TLEA-CMA-ES follow different sequences for optimizing multiple lower-level tasks.
BL-CMA-ES performs complete optimization for each lower-level task sequentially, while TLEA-CMA-ES performs only one lower-level iteration for each task in each parallel round, with successive parallel rounds continuing until all tasks meet the termination condition.
In contrast, DRC-CMA-ES optimizes tasks sequentially with one lower-level iteration only in the first parallel round.
In subsequent quasi-parallel rounds, tasks are selected based on the selection probabilities.

Accordingly, it is clear from Fig. \ref{fig:Resource_Allocation_compare}(g)-(i) that the opportunities for tasks to be optimized are unequal in DRC-CMA-ES.
Competitive tasks are assigned more executions, while tasks with low potential, such as task 3 in Fig. \ref{fig:Resource_Allocation_compare}(h), are executed only a few iterations.
Moreover, the selection probability not only ensures that every task has a chance to be selected, but also enables some tasks that exhibit a rapid improvement in performance to come from behind, such as task 1 in Fig. \ref{fig:Resource_Allocation_compare}(g).
In contrast, BL-CMA-ES and TLEA always perform complete optimization for each task, resulting in a significant resource waste.
For example, tasks like task 7 in Fig. \ref{fig:Resource_Allocation_compare}(b) and task 6 in Fig. \ref{fig:Resource_Allocation_compare}(d) consumed a large amount of lower-level FEs, but the resulting ($x_u$, $x_l^*$) pairs were not competitive enough and were eliminated from subsequent environment selection.
Instead, DRC-CMA-ES reduces such resource waste through effective resource allocation, thereby achieving a higher effective utilization rate of lower-level executions than the other algorithms.

\vspace{-0.5em}
\subsection{Running Time} 
In addition to the number of FEs, execution time also reflects the algorithm efficiency.
Fig. \ref{fig:Running_time} reports the average running time of different algorithms on SMD problems with dimension ($m$ = 30, $n$ = 30).
It is apparent that DRC-CMA-ES is faster than other algorithms on almost all the test problems. 
The advantages of DRC-CMA-ES are particularly pronounced on difficult problems such as SMD7 to SMD12.
By eliminating redundant lower-level iterations, DRC-CMA-ES not only reduces the number of FEs but also significantly improves overall speed.

\vspace{-0.5em}
\subsection{Real-world BLOPs Study} 
In this subsection, the algorithms are applied to two real-world BLOPs.
The first one is the gold mining problem in Kuusamo (denoted as GM):
the government at the upper level sets tax rates to maximize the revenue from mining project and minimize the pollution, while the mining company at the lower level controls the mining production accordingly to maximize its profits and minimize the reputation damage.
The second one is the decision making problem (denoted as DM):
the CEO at the upper level tries to maximize the profit and the quality of the products, while the division heads at the lower level intends to maximize branch profits and the workers’ satisfaction.
Details of these problems can be found in \cite{islam2017enhanced} and the scalarization method for handling multiple objectives follows \cite{he2018evolutionary}.

The results of function values and FEs on GM and DM are summarized in Table \ref{table:GM and DM}.
It can be observed that DRC-CMA-ES obtains the best lower-level function values on both problems, as well as the best upper-level function values on gold mining problem.
Meanwhile, DRC-CMA-ES consumes significantly less computing resources compared to other algorithms.

\section{Conclusion} 
The aim of this research is to reduce the substantial resource consumption in evolutionary bilevel optimization.
The proposed method successfully achieves this goal by eliminating redundant lower-level optimizations.
Computing resources are prioritized to promising lower-level tasks through a competitive quasi-parallel framework.
The execution opportunities of tasks are determined by a comprehensive selection probability, which not only ensures that highly competitive tasks are prioritized, but also allows promising tasks to come from behind. 
Additionally, a cooperative knowledge transfer mechanism based on evolution information and navigational individual is embedded within the competition, which further improves efficiency and reduces the risk of falling into local optima.

Experimental results on test suite instances and two real-world problems demonstrate that the proposed DRC-BLEA significantly reduces the resource consumption while achieving competitive accuracy across various problem scales.
Moreover, with resources fully utilized, DRC-BLEA is obviously faster than existing methods.
We compared the resource allocation behavior of DRC-BLEA with other algorithms, and analyzed the reasons behind the performance improvement of the proposed framework.
For future work, we plan to introduce predictive machine learning models to guide the resource allocation.
In addition, extending DRC-BLEA to bilevel multi-objective optimization scenarios is a promising direction.

\ifCLASSOPTIONcaptionsoff
  \newpage
\fi

\footnotesize
\bibliographystyle{IEEEtran}
\bibliography{mybibtex}

\end{document}